%% file: main.tex
\documentclass[10pt,twocolumn,letterpaper]{article}

\usepackage[pagenumbers]{cvpr} % To force page numbers, e.g. for an arXiv version

\usepackage{graphicx}
\usepackage{amsmath}
\usepackage{amssymb}
\usepackage{booktabs}
\usepackage{enumitem}
\usepackage{times}
\usepackage{epsfig}
\usepackage{graphicx}
\usepackage{makecell}

\input{preamble}

\definecolor{cvprblue}{rgb}{0.21,0.49,0.74}
\usepackage[pagebackref,breaklinks,colorlinks,citecolor=cvprblue]{hyperref}
\usepackage{multirow}

\def\paperID{12619} % *** Enter the Paper ID here
\def\confName{CVPR}
\def\confYear{2024}

\def\method{\text{MixPL}}
\title{Mixed Pseudo Labels for Semi-Supervised Object Detection}

\author{
Zeming Chen$^{1}$\thanks{This work is done when Zeming Chen was intern in Shanghai AI Lab.}
\quad Wenwei Zhang$^{2,4}$
\quad Xinjiang Wang$^3$
\quad Kai Chen$^4$
\quad Zhi Wang$^1$ 
\\
{$^1$Shenzhen International Graduate School, Tsinghua University} \\
\quad {$^2$S-Lab, Nanyang Technological University} 
\quad {$^3$SenseTime Research} 
\quad {$^4$Shanghai AI Laboratory} \\
\quad \small{\texttt{czm20@mails.tsinghua.edu.cn}} 
\quad \small{\texttt{wenwei001@ntu.edu.sg}} 
\quad \small{\texttt{wangxinjiang@sensetime.com}} \\
\quad \small{\texttt{chenkai@pjlab.org.cn}}
\quad \small{\texttt{wangzhi@sz.tsinghua.edu.cn}}
}

\begin{document}
\maketitle
\input{sec/0_abstract}    
\input{sec/1_intro}
\input{sec/2_relate}
\input{sec/3_method}
\input{sec/4_analysis}
\input{sec/5_exp}
\input{sec/6_con}

{
    \small
    \bibliographystyle{ieeenat_fullname}
    \bibliography{main}
}

\clearpage
\onecolumn
\appendix
%\counterwithin{figure}{section}
\renewcommand{\thefigure}{A\arabic{figure}}
\setcounter{table}{0}
\renewcommand{\thetable}{A\arabic{table}}

\input{sec/7_suppl}


\end{document}

%% file: preamble.tex
\usepackage[dvipsnames]{xcolor}

%% file: sec/0_abstract.tex
\begin{abstract}
While the pseudo-label method has demonstrated considerable success in semi-supervised object detection tasks, this paper uncovers notable limitations within this approach. Specifically, the pseudo-label method tends to amplify the inherent strengths of the detector while accentuating its weaknesses, which is manifested in the missed detection of pseudo-labels, particularly for small and tail category objects.
To overcome these challenges, this paper proposes Mixed Pseudo Labels (\method), consisting of Mixup and Mosaic for pseudo-labeled data, to mitigate the negative impact of missed detections and balance the model's learning across different object scales. Additionally, the model's detection performance on tail categories is improved by resampling labeled data with relevant instances. 
Notably, \method~consistently improves the performance of various detectors and obtains new state-of-the-art results with Faster R-CNN, FCOS, and DINO on COCO-Standard and COCO-Full benchmarks. 
Furthermore, \method~also exhibits good scalability on large models, improving DINO Swin-L by 2.5\% mAP and achieving nontrivial new records (60.2\% mAP) on the COCO \texttt{val2017} benchmark without extra annotations.
Codes are available at~\url{https://github.com/Czm369/MixPL}.
\end{abstract}

%% file: sec/1_intro.tex
% !TEX root = ../main.tex
\section{Introduction}\label{sec:Introduction}
Object detection has seen remarkable advancements~\cite{ren2015faster, chen2019htc, detr, zhu2020deformable} in the deep learning era, but heavily relies on costly human annotations.
Therefore, semi-supervised learning~\cite{lee2013pseudo, sohn2020fixmatch, xu2021end} has been attracting more and more research interests, which exploits unlabeled data to improve detector performance without solely relying on annotations.

Semi-supervised learning approaches~\cite{tarvainen2017mean, lee2013pseudo} involves generating pseudo-labels from model predictions for unlabeled data and using them to train a student model.
Such a paradigm has been widely successful in image classification~\cite{sohn2020fixmatch, wang2022freematch, zheng2022simmatch} but faces significant challenges in object detection.
As object detectors are required to detect objects of varying sizes and numbers within an image, 
their predictions, later used as pseudo-labels, may either miss objects or contain bounding boxes of background regions.
Additionally, detectors have different capabilities in detecting objects of different scales and categories, leading to a significant difference in the distribution of objects between pseudo labels and ground truth. 
Thus, pseudo-labels foster notable enhancements in recognizing large objects and prominent categories, yet offer limited benefits for identifying small objects and tail categories.

\begin{figure}[t] 
    \vspace{-6pt}
    \includegraphics[width=0.9\linewidth]{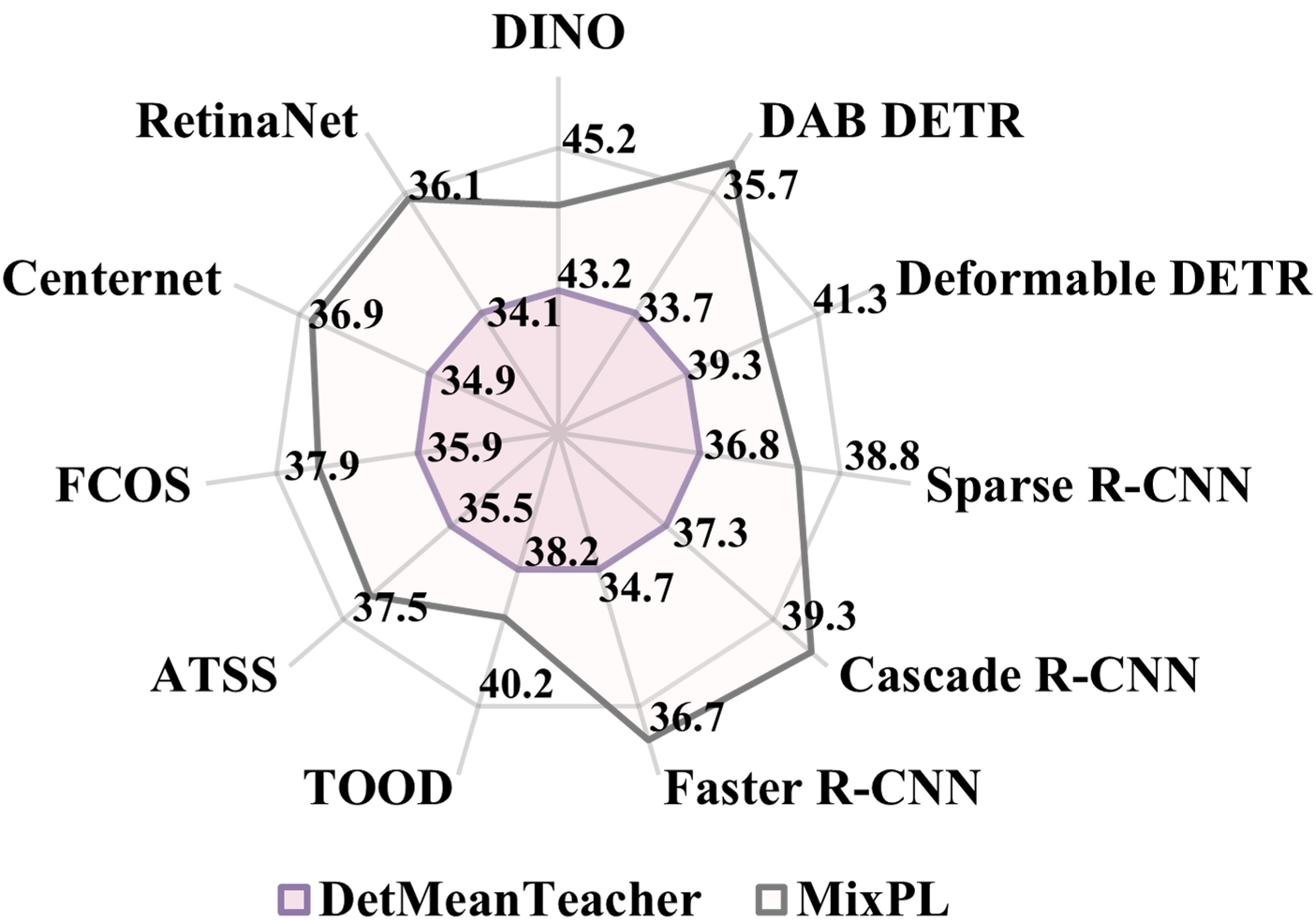}
    \centering
    \caption{The performances of DetMeanTeacher and \method~on COCO 10\%. 
             Based on DetMeanTeacher, \method~unanimously improves different kinds of object detectors and obtains new state-of-the-art performance.}
    \label{fig:performance_coco-10}
    \vspace{-6pt}
\end{figure}

Previous methods in semi-supervised object detection (SSOD) improved pseudo-label quality~\cite{xu2021end} or learning efficiency~\cite{chen2022dense, li2022pseco, liu2021unbiased} by enhancing specific components of detectors. However, these approaches lacked universality across different models. In contrast, our method universally enhances model performance by tailoring adjustments to different scales and categories, leveraging differences between pseudo-labels and ground truth.

\begin{figure}[h]
    \vspace{-6pt}
    \centering
    \includegraphics[width=0.9\linewidth]{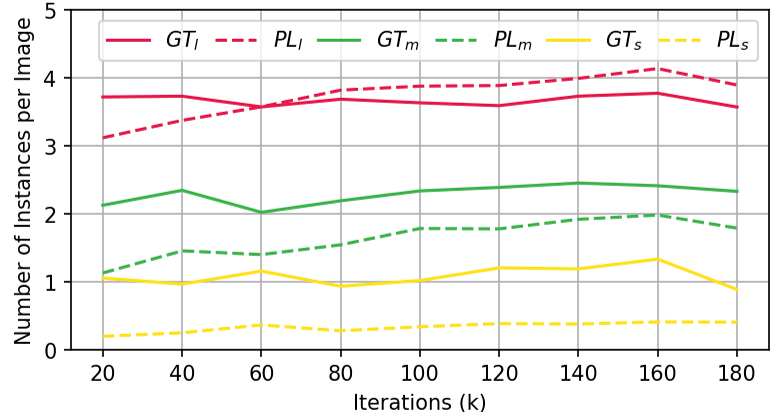}
    \caption{Scale distribution of ground truth and pseudo-labels for Faster R-CNN on COCO 10\%. 
    $GT_{l/m/s}$ represents the number of large/medium/small objects in ground truth, 
    and $PL_{l/m/s}$ represents the number of large/medium/small objects filtered by the threshold in pseudo-labels. 
    The $PL_l$ gradually exceeds $GT_l$, but the $PL_m$ and $PL_s$ is always lower than $GT_m$ and $GT_s$.}
    \label{fig:num_per_image}
    \vspace{-6pt}
\end{figure}

\begin{figure}[h]
    \vspace{-6pt}
    \centering
    \includegraphics[width=0.9\linewidth]{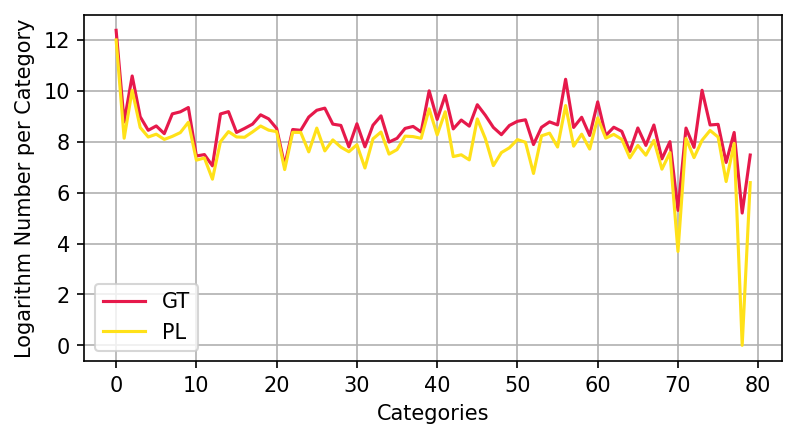}
    \caption{Category distribution of ground truth and pseudo-labels for Faster R-CNN on COCO 10\%. 
    $GT$ and $PL$ represent the logarithm number of ground truth and pseudo-labels, respectively.}
    \label{fig:lognum_per_cat}
    \vspace{-6pt}
\end{figure}

To this end, we first summarize a detection MeanTeacher (DetMeanTeacher) framework.
Similar to MeanTeacher~\cite{tarvainen2017mean} for image classification, DetMeanTeacher is conceptually simple and applicable to \emph{all} kinds of object detectors.
Although MeanTeacher has been used as a basic framework in previous SSOD methods~\cite{xu2021end, liu2021unbiased, chen2022dense},
we found it non-trivial to make DetMeanTeacher work on different kinds of object detectors, especially on one-stage detectors~\cite{lin2017focal, tian2019fcos}. These detectors necessitate filtering out images with no pseudo-labels due to significant deviations from ground truth. This inspires us to leverage the DetMeanTeacher to measure differences between pseudo-labels and ground truth in quantity, scale, and category.
As shown in Fig.~\ref{fig:num_per_image} and Fig.~\ref{fig:lognum_per_cat}, we conduct a statistical analysis on pseudo-labels generated by Faster R-CNN based on DetMeanTeacher and propose Mixed Pseudo Labels (\method) with Labeled Resampling to
improve the semi-supervised performance of detectors.

\begin{figure}[t]
    \vspace{-6pt}
    \includegraphics[width=0.9\linewidth]{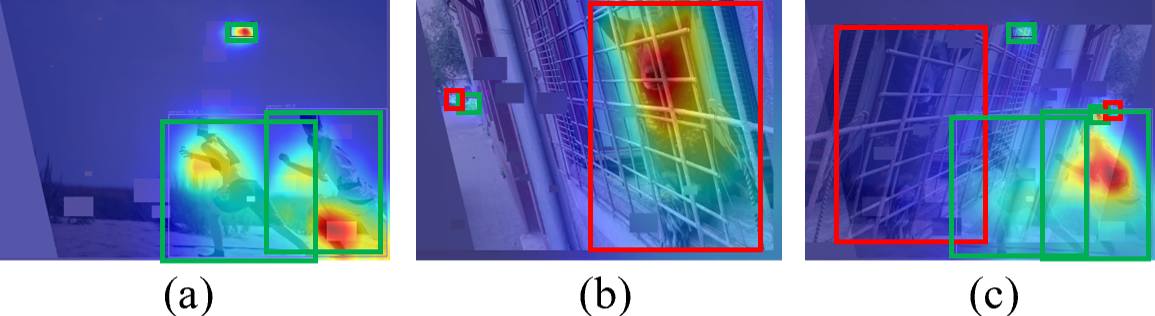}
    \centering
    \caption{Grad-CAM of different augmented images with pseudo-labels for Faster R-CNN on COCO 10\%. 
    (c) is the Pseudo Mixup image of (a) and (b). 
    The green boxes indicate correct pseudo-labels and the red boxes indicate missed objects. 
    Pseudo Mixup is effective in reducing the gradient response of missed objects and enhancing the gradient response of correct pseudo-labels.}
    \label{fig:gradnorm_strong_mixup}
    \vspace{-6pt}
\end{figure}

\textbf{Quantity.} 
During early training, the quantity of pseudo-labels is noticeably lower than the ground truth. 
In fact, due to the background-foreground sample imbalance, the model tends to label samples as background. 
Pseudo-labels amplify this by optimizing the missed foreground as background in unlabeled data, reinforcing the bias toward classifying samples as background.
Although missed detection is difficult to avoid, we propose Pseudo Mixup to reduce negative effects. This technique involves a pixel-wise overlay of two pseudo-labeled images, effectively mitigating the overall negative impact of false negative samples by making them more akin to actual negatives, as shown in Fig.~\ref{fig:gradnorm_strong_mixup}.

\textbf{Scale.} Throughout the training, the quantity of large objects in pseudo-labels gradually surpasses ground truth, while small and medium objects consistently lag behind ground truth, as shown in Fig.~\ref{fig:num_per_image}. 
The detectors struggle with smaller objects, and pseudo-labels exacerbate this by treating them as background in unlabeled data during optimization.
However, the model excels in detecting large objects, so we propose Pseudo Mosaic to transform large objects in pseudo-labels into small objects. Pseudo Mosaic combines 4 down-sampling pseudo-labeled images, providing more labels for small and medium objects. and enhancing detection capabilities across various scales.

\textbf{Categories.} After the training, pseudo-label quantities for head categories match ground truth, while certain tail categories face deficits, as shown in Fig.~\ref{fig:lognum_per_cat}. 
Similar to the scale issue, the detectors struggle with tail-category objects, and pseudo-labels tend to treat these objects in unlabeled data as background, even leading to detection degradation in certain tail categories. 
Unlike tackling the scale issue, we cannot convert pseudo-labels of head categories to tail categories.
Instead, we propose Labeled Resampling to oversample tail categories in labeled data. Although this may reduce head category samples, the model effectively extracts head category pseudo-labels from unlabeled data, ensuring performance across all categories.

We conduct extensive experiments on DetMeanTeacher and \method. As shown in Fig.~\ref{fig:performance_coco-10},
DetMeanTeacher unanimously works with different kinds of detectors on COCO 10\%.
\method~further boosts the performance of DetMeanTeacher across detectors and obtains new state-of-the-art results in Faster R-CNN, FCOS, and DINO on COCO-Standard and COCO-Full benchmarks.
In addition, combined with Swin-L~\cite{swin} and DINO~\cite{zhang2022dino}, \method~exhibits good scalability and achieves nontrivial records (60.2\% mAP) on COCO \texttt{val2017} without extra labeled data.

%% file: sec/2_relate.tex
% !TEX root = ../main.tex
\section{Related Works}
\subsection{Semi-supervised image classification}
Semi-supervised image classification aims to improve classifiers by utilizing unlabeled data in conjunction with labeled data.  
Semi-supervised learning techniques in image classification can be categorized into two paradigms: 
1) pseudo labels~\cite{lee2013pseudo, xie2020self} that uses model predictions on unlabeled data as labels to train the model,
and 2) consistency regularization~\cite{laine2016temporal, tarvainen2017mean, xie2020unsupervised} that imposes consistency among the model predictions of the same image.
Recent advances\cite{berthelot2019mixmatch, sohn2020fixmatch, zhang2021flexmatch, zheng2022simmatch, wang2022freematch} synergistically combine the two paradigms.
FixMatch, one of the representatives, predicts pseudo-labels on weakly augmented unlabeled images and then trains a model on strongly augmented pseudo-labeled images to ensure consistency between weak and strong augmentations.
Notably, these semi-supervised methods generally apply to different classifiers and are not customized for a particular architecture.
On the same merit, this paper aims to design an effective detector-agnostic scheme for semi-supervised object detection (SSOD).

\subsection{Object detection}
Object detection requires recognizing and localizing objects of interest in an image. 
Due to the scale, position, and number of objects vary across images, different detection architectures are proposed to tackle the challenges, including one-stage~\cite{lin2017focal, tian2019fcos, feng2021tood}, two-stage or cascaded detectors~\cite{ren2015faster, chen2019htc, cai2019cascade}, and detection transformers (DETR)~\cite{zhu2020deformable, zhang2022dino,liu2022dab,detr}. 
Orthogonal to detection architectures, different feature pyramid networks (FPN) ~\cite{lin2017fpn, nasfpn} are developed to enhance multi-scale features and detect objects across scales.
Significantly, this paper verifies the validity of \method~on detectors across different structures.

\subsection{Data augmentation}
Data augmentation is crucial in visual perception and semi-supervised learning.
Apart from photometric and geometric transformations like color jittering, resizing, and flipping~\cite{mmdetection, wu2019detectron2}, 
augmentations that mix training data become popular in semi-supervised learning and object detection, such as CutOut~\cite{devries2017improved}, Random Erasing~\cite{zhong2020random}, Mixup~\cite{zhang2017mixup}, and Mosaic~\cite{yolov4}.
While Mixup and Mosaic have broad applications~\cite{rtmdet, yolox, yolov4}, this paper unveils their novel roles in SSOD, thus overcoming the limitation of pseudo-labels.

\subsection{Semi-supervised object detection}
Different from classifiers that only have one error type in their predictions, there are various error types in the predictions of detectors due to the challenge of object detection, including missed objects, bounding boxes with wrong categories or inaccurate positions.
Consequently, besides directly adopting existing semi-supervised learning paradigms~\cite{jeong2019consistency, jeong2021interpolation, sohn2020simple}, recent methods
explore special designs for particular object detectors to improve the quality of pseudo-labels and explore better learning strategies~\cite{xu2021end, chen2022dense, li2022pseco, liu2021unbiased,  chen2022label, zhou2022dense, zhang2023semi}.
For example, SoftTeacher utilizes R-CNN to filter uncertain pseudo-labels, while cannot easily transfer to one-stage detectors.
Dense Teacher explores learning from dense labels in one-stage detectors, making it not directly applicable to two-stage detectors.
PseCo learns scale invariance by exploiting FPN, which might be absent in DETR.
Although obtaining promising results, these special designs may limit their applications to a broader set of detectors and hinder the understanding of SSOD from a unified perspective.
In contrast, this paper studies the common issues for pseudo-labels in SSOD without any assumptions on detector architectures.

\begin{figure*}[ht]
    \centering
    \includegraphics[width=0.9\linewidth]{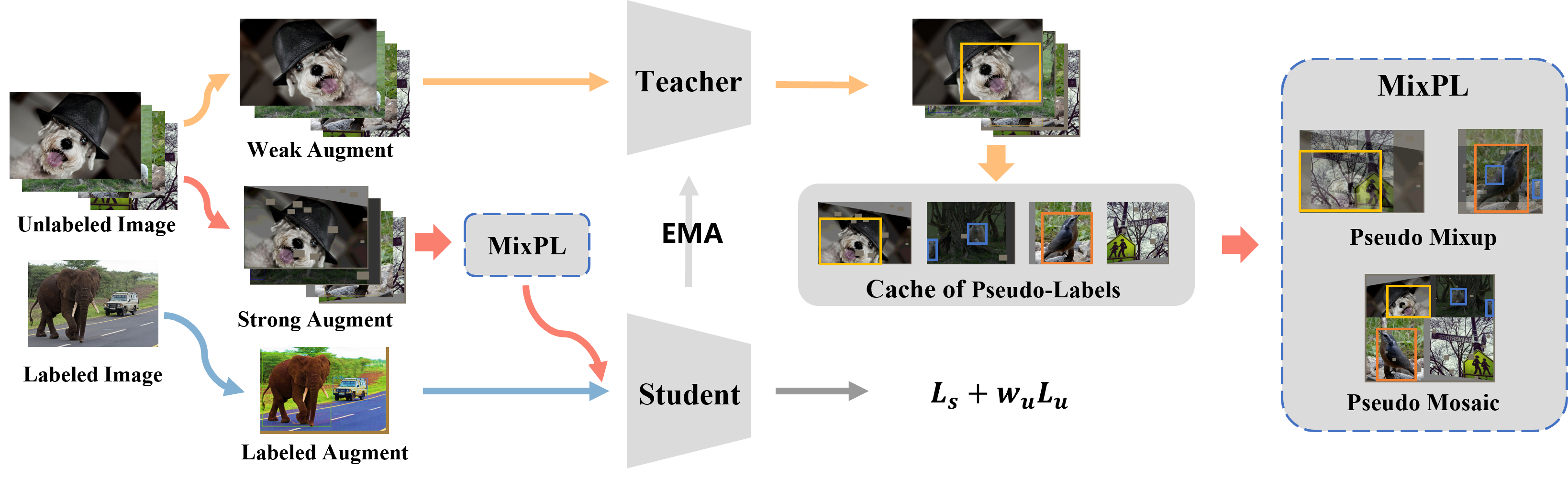}
    \vspace{-6pt}
    \caption{The framework of DetMeanTeacher and \method.}
    \label{fig:meanteacher}
    \vspace{-12pt}
\end{figure*}

%% file: sec/3_method.tex
% !TEX root = ../main.tex
% \section{Methodology}\label{sec:methods}
\section{Detection MeanTeacher}\label{sec:meanteacher}
\begin{table*}[h]
\begin{minipage}[t]{0.31\textwidth}
\scalebox{0.6}{\input{tables/meanteacher_stable}}
\centering
% \vspace{-6pt}
\subcaption{Effectiveness of filtering \emph{empty images}.}
\label{tab:meanteacher_stable}
\end{minipage}
\hspace{0.01\textwidth}
\begin{minipage}[t]{0.31\textwidth}
\scalebox{0.6}{\input{tables/meanteacher_thr}}
\centering
% \vspace{-6pt}
\subcaption{Impact of confidence threshold.}
\label{tab:meanteacher_thr}
\end{minipage}
\hspace{0.01\textwidth}
\begin{minipage}[t]{0.31\textwidth}
\scalebox{0.6}{\input{tables/meanteacher_weight}}
\centering
% \vspace{-6pt}
\subcaption{Impact of loss weight $w_u$.}
\label{tab:meanteacher_weight}
\end{minipage}
\vspace{-6pt}
\centering
\caption{Analysis of DetMeanTeacher.}
\end{table*}
In this study, we aim to explore effective methods that are applicable to different object detectors for SSOD.
Therefore, this paper starts with the analysis of MeanTeacher framework, which is the core framework shared among recent SSOD methods.
As shown in Fig.~\ref{fig:meanteacher}, MeanTeacher framework contains a teacher model and a student model that have the same structure.
During each training iteration: 
\begin{itemize}
  \item[1.]
  Labeled and unlabeled images are randomly sampled to form a batch following a given ratio.
  \item[2.]
  Each unlabeled image is augmented by weak and strong augmentations to obtain two views, and the teacher predicts pseudo-labels on weakly augmented images.
  \item[3.]
  The pseudo-labels are filtered by a confidence threshold and transformed to the strong view to form pseudo-labeled data with strongly augmented images.
  \item[4.]
  The student is updated by optimizing supervised loss $L_s$ and unsupervised loss $L_u$ based on labeled and pseudo-labeled data, and the teacher is updated by using the exponential moving average (EMA) of the student.
\end{itemize}

Theoretically, MeanTeacher framework does not assume a specific detector structure and thus should apply to all kinds of detectors.
To verify that and understand the impact of MeanTeacher on different detectors, we take RetinaNet and Faster R-CNN as the representative of one-stage and two-stage detectors, respectively, and then apply MeanTeacher to them for semi-supervised learning.
The analysis is conducted on the COCO dataset with 10\% labeled data and with the rest 90\% as unlabeled data.
The supervised baseline of Faster R-CNN and RetinaNet on the COCO 10\% labeled data is 26.6\% AP and 27.2\% AP, respectively.
The confidence threshold used in Faster R-CNN and RetinaNet to filter pseudo-labels is 0.7 and 0.4, respectively.

\textbf{Filtering images with no pseudo labels.}
As shown in Table~\ref{tab:meanteacher_stable}, Faster R-CNN can obtain a reasonable improvement on its supervised baseline (34.7\% vs. 26.6\% AP).
In contrast, RetinaNet cannot even surpass its supervised baseline and only achieves 17.9\% AP.
We empirically find that some images (called \emph{empty images} for short) actually do not contain any pseudo-labels when training RetinaNet.
This is because the teacher model is not perfect, some of its predicted bounding boxes will have low confidence and will be filtered by the confidence threshold.
As the disparity between positive and negative samples is an inherent issue of one-stage detectors, such an issue can be significantly exacerbated by these \emph{empty images}, making training unstable.
To solve this issue, we filter these \emph{empty images} in training, i.e., the student model will not learn on these \emph{empty images}.
This simple strategy brings 90.5\% relative improvement of RetinaNet, achieving 34.1\% AP.

\textbf{Balance precision and recall of pseudo labels.}
We investigate the impact of confidence threshold with the strategy of filtering \emph{empty images} to stabilize training.
As shown in Table~\ref{tab:meanteacher_thr}, both RetinaNet and Faster R-CNN have their respective optimal confidence thresholds, a deviation from which leads to an obvious performance drop.
The results indicate that a suitable confidence threshold is necessary for the effectiveness of MeanTeacher and is related to the specific classification loss function.

\textbf{Balance learning on labeled and unlabeled data.}
We investigate the impact of $w_u$ in MeanTeacher.
It is observed that the optimal $w_u$ for both RetinaNet and Faster R-CNN is $2$ instead of $4$ used in previous studies ~\cite{xu2021end} when \emph{empty images} filtering and a lower confidence threshold are adopted. 

The proposed modification results in a general framework for SSOD, termed Detection MeanTeacher (DetMeanTeacher), with simple strategies to reduce the impact of \emph{empty images} and the side effects of pseudo-labels.
Being simple yet effective, DetMeanTeacher yields competitive performance in comparison with existing SSOD methods that adopt more sophisticated designs.
Therefore, we decouple detectors from the semi-supervised learning framework, allowing us to observe the consistency in differences between pseudo-labels and ground truth across various detectors based on DetMeanTeacher. This insight inspires the proposal of \method~with Labeled Resampling.

\section{Mixed Pseudo Labels}\label{sec:mixpl}
The investigations of DetMeanTeacher prompt us to consider missed detection in pseudo-labels, since \emph{empty images} can be viewed as extreme cases of missed detection, implying the omission of all objects. 
Thus, the confidence threshold reflects the degree of missed detection to some extent, and the $w_u$ aims to ensure the interference of missed detection is less than the signal of supervised learning.
The inherent cause of missed detection is the foreground-background imbalance, leading the model to be inclined to classify samples as background.
Simultaneously, due to the model's inadequacy in detecting small objects and tail-category objects, missed detection is concentrated on these elements, as depicted in Fig.~\ref{fig:num_per_image} and Fig.~\ref{fig:lognum_per_cat}.
Thus, we propose Mixed Pseudo Labels (\method), consisting of Pseudo Mixup and Pseudo Mosaic that eliminate the negative effects of false negative samples and supplement the training samples of small and medium objects. 
In addition, we also propose that Label Resampling helps MixPL significantly improve the detection performance of tail categories.

\textbf{Pseudo Mixup.}
As analyzed in DetMeanTeacher, the confidence threshold filters predicted bounding boxes of low confidence, which tends to make the pseudo-labels miss some objects of interest and causes continuous interference to the training of the model when the positive samples are optimized as negative samples.

Considering that the missed detection is attributed to the imbalance between foreground and background samples, we proposed Pseudo Mixup. We leverage this imbalance to overlay two pseudo-labeled images and unite their pseudo-labels, ensuring that a positive sample is more likely to be superimposed with a negative sample. Then, the missed positive samples become more like the negative samples because negative samples are superimposed, so the overall negative effect of this part of samples will be significantly reduced.
Additionally, we discover that Pseudo Mixup acts as an effective means of introducing robust perturbations for accurate positive samples and amplifying their gradient response, which proves beneficial for model training.

\begin{figure*}[t]
    \includegraphics[width=1.0\linewidth]{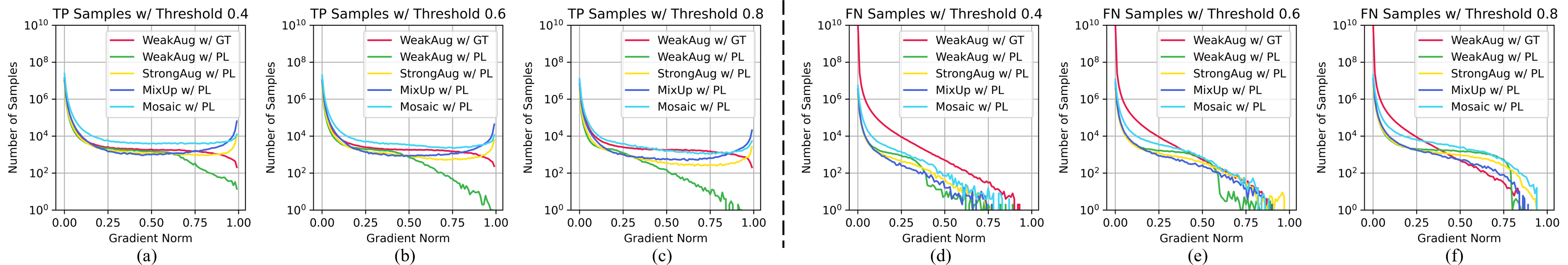}
    \centering
    \caption{Gradient density for different augmentations with ground truth (GT) or pseudo labels (PL). 
    (a)-(c) represent gradient density of TP samples at various confidence thresholds; 
    (d)-(f) represent gradient density of FN samples at various confidence thresholds.}
    \label{fig:gradnorm_mixup}
\end{figure*}

\textbf{Pseudo Mosaic.}
A closer look at Table~\ref{tab:meanteacher_thr} reveals that AP${_s}$ of Faster R-CNN consistently decreases when increasing the threshold, while AP${_m}$ and AP${_l}$ of RetinaNet consistently increases along with an increasing threshold.
Such a trend implies a strong correlation between the sizes and the confidence of the predicted bounding boxes.
In Table~\ref{tab:size_ratio}, we have computed the mean scores of bounding boxes with varying sizes. It is observed that the average scores of bounding boxes increase as their sizes increase. 
Such a correlation eventually makes the distribution of object sizes in pseudo-labels significantly deviate from ground truth.

\begin{table}[h]
\vspace{-1.0em}
% \vspace{-9pt}
\scalebox{0.9}{\input{tables/size_ratio}}
% \vspace{-9pt}
\vspace{-6pt}
\centering
\caption{Average scores of bounding boxes of different sizes.}
\label{tab:size_ratio}
% \vspace{-9pt}
\vspace{-1.0em}
\end{table}

To alleviate this issue, we propose Pseudo Mosaic.
Specifically, Pseudo Mosaic takes 4 pseudo-labeled images and down-samples them, then composes these images into 1 image, which contains all the pseudo-labels in the original images.
It not only increases the number of pseudo-labels in the image by 4 times but also creates more small and medium objects down-sampled from their original images.
As a result, the distribution of object sizes in the pseudo-labels is much better balanced.

\textbf{Labeled Resampling.}
The detectors exhibit a significant performance disparity in detecting head and tail categories, resulting in an inadequate number of instances for certain tail categories in pseudo-labels. Different from scale problems, converting head categories into tail categories is unfeasible. Thus, we propose Labeled Resampling in combination with the semi-supervised learning scenario. The oversampling of tail categories from labeled data significantly improves the accuracy of tail categories and detectors mine enough head category pseudo-labels from the unlabeled data to balance the under-sampling of head categories in labeled data, which improves the model performance across all categories. The process is as follows.

First, compute the fraction of labeled data containing each category $c$: $f(c)$.
Then, compute the category-level repeat factor $r(c) = 1 / f(c)^{power}$, $power$ is a hyper-parameter with values in [0, 1]. 
Finally, compute the image-level repeat factor for each image $I$: $r(I)=\mathrm{max}_{c \in I}r(c)$.

\textbf{Implementations.}
As shown in Fig.~\ref{fig:meanteacher}, \method~is applied after obtaining the strongly augmented unlabeled images in DetMeanTeacher.
Since \method~requires more pseudo-labeled images to mix, during each iteration, the unlabeled images in the current batch are mixed with those sampled from the cache of pseudo-labels, which stores pseudo-labels and their corresponding images in the nearest previous iteration. 
% The update of the cache follows the first-in-first-out manner. 
For example, given 1 labeled image and 4 unlabeled images at each training iteration, we first sample 4 more images with their pseudo-labels from the cache and obtain 4 mixed images by Pseudo Mixup. Then we form 1 mixed image using Pseudo Mosaic that randomly combines 4 of the 8 images. These 5 mixed images are used to replace the original 4 unlabeled images and train the student model.

%% file: tables/meanteacher_stable.tex
% \tablestyle{2pt}{1.0}
\scalebox{0.9}{
\centering
\begin{tabular}{c|c |c |c c|c c c}
\toprule
Detector & Filter & AP & AP$_{50}$ & AP$_{75}$ & AP$_{s}$  & AP$_{m}$  & AP$_{l}$ \\
\midrule
\multirow{2}{*}{Faster R-CNN} 
& $\times$ & 34.7 & 54.6 & 37.6 & 19.5 & 37.1 & 46.1\\
& \checkmark & 34.7 & 54.7 & 37.4 & 19.3 & 37.6 & 45.8\\
\midrule
\multirow{2}{*}{RetinaNet}
& $\times$ & 17.9 & 29.0 & 18.8  & 8.1 & 21.7 & 29.6 \\
& \checkmark & 34.1 & 52.7 & 36.0 & 18.1 & 36.8 & 46.2\\
\bottomrule
\end{tabular}
}

%% file: tables/meanteacher_thr.tex
 % \tablestyle{3pt}{1.0}
  \scalebox{0.9}{

\begin{tabular}{c | c |c |c c|c c c}
\toprule
Detector & Thr & AP & AP$_{50}$ & AP$_{75}$ & AP$_{s}$  & AP$_{m}$  & AP$_{l}$ \\
\midrule
\multirow{3}{*}{Faster R-CNN} 
& 0.5 & 33.8 & 53.8 & 36.0 & 19.5 & 36.4 & 44.5 \\
& 0.7 & 34.7 & 54.7 & 37.4 & 19.3 & 37.6 & 45.8 \\
& 0.9 & 33.6 & 52.8 & 36.3 & 18.2 & 36.3 & 45.0 \\
\midrule
\multirow{3}{*}{RetinaNet}   
& 0.3 & 29.0 & 46.7 & 30.6 & 15.0 & 31.2 & 39.4 \\
& 0.4 & 34.1 & 52.7 & 36.0 & 18.1 & 36.8 & 46.2 \\
& 0.5 & 33.9 & 51.6 & 36.1 & 17.2 & 37.1 & 47.0 \\
\bottomrule
\end{tabular}
}

%% file: tables/meanteacher_weight.tex
 % \tablestyle{2pt}{1.0}
  \scalebox{0.9}{

\begin{tabular}{c| c |c |c c|c c c}
\toprule
Detector & $w_u$ & AP & AP$_{50}$ & AP$_{75}$ & AP$_{s}$  & AP$_{m}$  & AP$_{l}$ \\
\midrule
\multirow{3}{*}{Faster R-CNN} 
& 1 & 33.3 & 53.2 & 35.9 & 18.0 & 35.5 & 43.9 \\
& 2 & 34.7 & 54.7 & 37.4 & 19.3 & 37.6 & 45.8 \\
& 4 & 34.2 & 54.0 & 36.8 & 18.1 & 36.8 & 45.4 \\
\midrule
\multirow{3}{*}{RetinaNet}  
& 1 & 33.6 & 52.1 & 35.5 & 18.3 & 36.3 & 45.5 \\
& 2 & 34.1 & 52.7 & 36.0 & 18.1 & 36.8 & 46.2 \\
& 4 & 29.7 & 47.2 & 31.3 & 15.2 & 31.1 & 41.4 \\
\bottomrule
\end{tabular}
}

%% file: tables/size_ratio.tex
 % \tablestyle{4pt}{1.0}
  \scalebox{0.9}{

\begin{tabular}{c | c |c |c}
\toprule
Detector & Small  & Meduim & Large \\
\midrule
Faster R-CNN    & 0.304  & 0.406 & 0.509\\
RetinaNet    & 0.147 & 0.186 & 0.205\\
\bottomrule
\end{tabular}
}

%% file: sec/4_analysis.tex
% !TEX root = ../main.tex

\section{Analysis}
We analyzed MixPL from the perspective of gradient and regularization to reveal the effectiveness of the method.

\textbf{Gradient.}
We analyze the impact of \method~in comparison with weak and strong augmentations by analyzing the \emph{gradient density} of samples~\cite{GHMLoss}.
Specifically, given pseudo-labeled images filtered by a threshold, pseudo-labeled images are augmented by weak augmentation, strong augmentation, Pseudo Mixup, or Pseudo Mosaic.
Then we feed the augmented pseudo-labels to the student model of RetinaNet and then calculate the \emph{gradient density} of samples.
The gradients come from the following 4 types of samples based on their assignment results by pseudo-labels and ground truth, i.e., True Positive (TP), False Positive (FP), True Negative (TN), and False Negative (FN).

TP and TN indicate that the assignment of samples based on pseudo-labels and ground truth is consistent, whose gradient is clean and desirable for the student model. 
On the contrary, the samples of FP and FN are caused by the incorrect pseudo-labels and will contribute gradients leading to a wrong optimization direction.
We analyze the impact of \emph{gradient density} of TP and FN, 
which implies the impact of correct and missed objects in pseudo-labels.

We first analyze the effectiveness of Pseudo Mixup by comparing the trend of these augmentations when the threshold value increases in Fig.~\ref{fig:gradnorm_mixup}.
The number of positive samples in pseudo-labels gradually decreases with the increased threshold, 
making the \emph{gradient density} of TP gradually decrease.
In such a situation, the \emph{gradient density} of higher norm significantly decreases if weak augmentation is used, indicating weak augmentation is not sufficient as strong augmentation for training the student model.
On the contrary, the \emph{gradient density} of Pseudo Mixup is higher than that of strong augmentation, especially in the regions of high norm and high threshold.
This trend verifies the effectiveness of Pseudo Mixup in providing more positive samples and its superiority in providing hard positive examples over strong augmentation.
Foremost, because interpolating two images eventually makes the objects in the mixed image hard to detect, 
Pseudo Mixup not only provides hard positive examples in TP but also diminishes the effect of missing predictions (FN).
The \emph{gradient density} of Pseudo Mixup from FN is significantly lower than those of strong augmentation as the number of missed objects grows,
which implies the superiority of Pseudo Mixup in reducing the side effects of missed objects in pseudo-labels.

\textbf{Regularization.} 
\method~mixes pseudo-labeled images in the current iteration with those stored in the cache, which implicitly makes the student model see different views of the same pseudo-labeled image in a short window of training iterations.
Such a behavior imposes an implicit consistency regularization for the student model~\cite{jeong2019consistency}.
We verify the hypothesis by applying another strong augmentation to the same batch of pseudo-labels in the training iteration and double the batch size of unlabeled data, i.e., letting the student model see two views of each pseudo-labeled image, noted as \emph{double view}.
As shown in Table~\ref{tab:analysis_consistency}, \emph{Double view} brings 0.6\%AP improvement over the baseline.
In this regard, \method~also benefits from view consistency as both Pseudo Mixup and Pseudo Mosaic leverage image cache in past iterations to create multiple views for consistency. It should be noted that \method~is more effective and efficient than \emph{Double View} as seen from its higher performance and smaller batch size during training.
We hypothesize the extra performance gain is because \method~provides a more balanced pseudo-labeled distribution and more effective gradients, as described in the previous paragraphs.
\begin{table}[h]
\vspace{-1.0em}
\scalebox{0.9}{\input{tables/analysis_consistency}}
\vspace{-6pt}
\centering
\caption{Comparison between \method~and \emph{Double View}.}
\label{tab:analysis_consistency}
\vspace{-1.0em}
\end{table}

%% file: tables/analysis_consistency.tex
 % \tablestyle{3pt}{1.0}
  \scalebox{0.9}{

\begin{tabular}{c | c |c |c |c }
\toprule
& Baseline & \emph{Double View} & Pseudo Mixup & \method \\
\midrule
AP  & 34.7     &   35.3     &     35.7    &  37.2      \\
\bottomrule
\end{tabular}
}

%% file: sec/5_exp.tex
% !TEX root = ../main.tex

\begin{table*}[h]
\vspace{-6pt}
\scalebox{0.9}{\input{tables/coco_standard}}
\vspace{-6pt}
\centering
\caption{Benchmark results for Faster R-CNN on COCO.}
\label{tab:coco_standard}
\end{table*}

\section{Experiments}\label{sec:Experiments}
\subsection{Dataset and Experiment Settings} 

\textbf{Datasets.} We evaluate our methods on MS-COCO~\cite{coco} and PASCAL VOC~\cite{everingham2010pascal}.
MS-COCO contains 118k, 123k, and 5k images in \texttt{trian2017}, \texttt{unlabeled2017}, and \texttt{val2017}, respectively. 
Pascal VOC includes both VOC2007 and VOC2012, with a total of 5011 images in \texttt{trainval} of VOC2007, 4952 images in the test set of VOC2007, and 11540 images in \texttt{trainval} of VOC2012. We study and evaluate our approach using the following 3 experimental settings:
\begin{itemize}
  \item [1.]
  \textbf{COCO-Standard:} We sample 1\%/2\%/5\%/10\% images from \texttt{train2017} randomly as labeled data and treat the rest as unlabeled data.
  \item [2.]
  \textbf{COCO-Full:} We use \texttt{train2017} as labeled data and \texttt{unlabel2017} as unlabeled data. 
  \item [3.]
  \textbf{VOC-Mixture:} We use \texttt{VOC2007 trainval} as labeled data and \texttt{VOC2012 trainval} as unlabeled data. 
\end{itemize}

The COCO benchmarks evaluate the trained models on \texttt{val2017} dataset, while the VOC benchmarks assess them on \texttt{VOC2007} test set.

\textbf{Implementation details.} For experiments on COCO-Standard, all detectors are trained on 8 GPUs with 5 images per GPU (1 labeled and 4 unlabeled images).
For experiments on COCO-Full and VOC-Mixture, all detectors are trained on 8 GPUs with 8 images per GPU (4 labeled and 4 unlabeled images). 
The detectors are trained by 180k iterations (except DINO is trained for 90k iterations to avoid overfitting) and 36k iterations in experiments on COCO-Standard and VOC-Mixture, respectively. 
In experiments of COCO-Full, Faster R-CNN and FCOS are trained by 540k iterations, with the learning rate decayed by 10 after the 360k iterations, while DINO is trained by 270k iterations without the learning rate decay.
We use SGD optimizer for all detectors with a constant learning rate of 0.01, momentum of 0.9, and weight decay of 0.0001, except for detection transformers~\cite {liu2022dab, zhu2020deformable, zhang2022dino} whose original optimizer and hyper-parameters are followed in this study.
\begin{table}[t]
\vspace{-6pt}
\scalebox{0.9}{\input{tables/coco_fcos}}
\vspace{-6pt}
\centering
\caption{Benchmark results for FCOS on COCO.}
\label{tab:coco_fcos}
\end{table}
\begin{table}[t]
\vspace{-6pt}
\scalebox{0.9}{\input{tables/coco_dino}}
\vspace{-6pt}
\centering
\caption{Benchmark results for DINO on COCO.}
\label{tab:coco_dino}
\end{table}
\begin{table}[t]
\vspace{-6pt}
\scalebox{0.9}{\input{tables/unlabel2017}}
\vspace{-6pt}
\centering
\caption{Results with Swin-L on COCO-Full.}
\label{tab:unlabel2017}
\end{table}
\subsection{Benchmark Results}
\textbf{COCO-Standard.}
When the labeled data varies from 1\% to 10\%, \method~consistently shows competitive results.
For Faster R-CNN, 
\method~obtains the second-best results on COCO 2\% and 
surpasses all previous methods on COCO 5\% and COCO 10\% in Table~\ref{tab:coco_standard}.
For FCOS and DINO,
\method~obtains the second-best results on COCO 5\% for FOCS and 
surpasses all previous methods under other settings in Table~\ref{tab:coco_fcos} and Table~\ref{tab:coco_dino}.

\textbf{COCO-Full.}
As shown in Table~\ref{tab:coco_standard}, Table~\ref{tab:coco_fcos} and Table~\ref{tab:coco_dino}, \method~consistently improves supervised Faster R-CNN, FCOS and DINO by 5.2\% AP, 6.4\% AP and 4.3\% AP with unlabeled data, respectively, achieving new state-of-the-art performance.
Notably, it is more challenging to improve one-stage detectors under the semi-supervised setting as one-stage detectors face more severe imbalance issues in positive and negative samples, while \method~significantly surpasses the previous best results of DenseTeacher by 1.5\% AP with FCOS, without involving any detector-specific designs.
In addition, \method~exhibits great scalability when applied to the cutting-edge object detector DINO with strong backbone Swin-L~\cite{swin} in Table~\ref{tab:unlabel2017}. 
Without bells and whistles, \method~surpasses previous best single model results achieved by SoftTeacher (60.2\%AP v.s. 59.9\% AP) without extra annotations in SSOD.
\begin{table}[t]
\vspace{-1.0em}
\scalebox{0.9}{\input{tables/voc0712}}
\vspace{-6pt}
\centering
\caption{Benchmark results on VOCC-Mixture.}
\label{tab:voc0712}
\vspace{-1.0em}
\end{table}

\textbf{VOC-Mixture.}
To further verify the generalization ability of \method~across different datasets, we compare different SSOD methods in Table~\ref{tab:voc0712}.
\method~is consistently better than previous methods, with either Faster R-CNN and FCOS, where the more remarkable advantages are observed on the one-stage detector FCOS.
This phenomenon is consistent with those observed on COCO-Full and further verifies the ability of \method~to alleviate the imbalance issue of one-stage detectors under semi-supervised learning.
\begin{table}[h]
\vspace{-1.0em}
\scalebox{0.9}{\input{tables/coco_10percent}}
\vspace{-6pt}
\centering
\caption{Improvements on various detectors on COCO 10\%.}
\label{tab:coco_10percent}
\vspace{-1.0em}
\end{table}

\textbf{Improvements on different detectors.}
We study the effectiveness of DetMeanTeacher and \method~across different object detectors on COCO 10\%, by conducting experiments on 11 object detectors, covering mainstream one-stage, two-stage, and end-to-end detectors.
The results in Table~\ref{tab:coco_10percent} reveal that \method~is a universal SSOD framework that is generally applicable to different kinds of object detectors.
Simply filtering \emph{empty images} with no pseudo labels predicted, DetMeanTeacher obtains at least 6.9\% AP improvement over all the supervised baselines.
In comparison with previous methods in Table~\ref{tab:coco_standard}, DetMeanTeacher already surpasses some existing SSOD methods that are specially designed for different detection architectures, e.g., Soft Teacher (34.7\%AP v.s. 34.04\% AP).
Moreover, \method~consistently improves DetMeanTeacher by at least 0.7\% AP across different object detectors.
The overall results also provide new data points for future research on universal SSOD.

\subsection{Ablation Study}
We conducted ablation experiments on Pseudo Mixup and Pseudo Mosaic based on DetMeanTeacher with Faster R-CNN. Both Pseudo Mixup and Pseudo Mosaic exhibit significant gains in Table~\ref{tab:ablation_overall}.
Meanwhile, Pseudo Mosaic remarkably improves the recognition ability of small and medium objects of the detector.

\textbf{Pseudo Mixup.} To verify the necessity of Pseudo Mixup, we explore mixing 1 pseudo-labeled image with 1 labeled image~\cite{tianfzhou2021instant}, which obtains lower performance than the baseline in Table~\ref{tab:ablation_mixup}.
Mixing 2 pseudo-labeled images yields better results and surpasses the baseline by 1\% AP.

\textbf{Pseudo Mosaic.} We then carefully ablate the effectiveness of image scales for Pseudo Mosaic in Table~\ref{tab:ablation_mosaic}. The resolution of 200 indicates the longest edge of each image to be composed will be 200. We first study down-sampling or up-sampling images to fixed sizes and find that 800 achieves the best AP among fixed sizes and 400 obtains the best AP$_s$. 
Then we study the suitable range of sizes to find a good trade-off between AP of different sizes and reduce the training cost.
The results show that 400$\sim$ 800 is a better trade-off than fixed sizes and other ranges of sizes.
\begin{table}[ht]
\vspace{-1.0em}
\scalebox{0.9}{\input{tables/ablation_overall}}
\vspace{-6pt}
\centering
\caption{Effectiveness of main components of~\method.}
\label{tab:ablation_overall}
\vspace{-1.0em}
\end{table}
\begin{table}[ht]
\vspace{-1.0em}
\scalebox{0.9}{\input{tables/ablation_mixup}}
\vspace{-6pt}
\centering
\caption{Exploring strategies of Mixup.}
\label{tab:ablation_mixup}
\vspace{-1.0em}
\end{table}
\begin{table}[ht]
\vspace{-1.0em}
\scalebox{0.9}{\input{tables/ablation_mosaic}}
\vspace{-6pt}
\centering
\caption{Resolutions of a single image for Pseudo Mosaic.}
\label{tab:ablation_mosaic}
\vspace{-1.0em}
\end{table}

\textbf{Labeled Resampling.} 
Due to limited labeled data sometimes leading to certain tail categories without labels, we only conduct ablation experiments on Faster R-CNN and FCOS configurations on COCO 10\%. Our analysis focuses on the performance of the most two classes (person and car) and the least two classes (toaster and hair drier), detailed in Table~\ref{tab:resampling_faster} and Table~\ref{tab:resampling_fcos}. Notably, the semi-supervised gains in head categories surpass those in tail categories for both Faster R-CNN and FCOS. This trend persists even when the original accuracy of head categories is already high, and in some cases, tail category performance falls below the supervised learning baseline.
Boosting labeled data with oversampled tail categories notably enhances "toaster" detection without a significant decline in "person" and "car." However, "hair dryer" detection shows limited improvement, highlighting the need for additional labeled data for challenging tail categories. 
Furthermore, Labeled Resampling shows a more significant improvement for FCOS compared to Faster R-CNN, primarily due to the fairness of Focal Loss in category-wise confidence, as shown in Table~\ref{tab:confidence_per_cat}.
\begin{table}[ht]
\vspace{-1.0em}
\scalebox{0.9}{\input{tables/labeled_resampling_faster}}
\vspace{-6pt}
\centering
\caption{Powers of Labeled Resampling on Faster R-CNN .}
\label{tab:resampling_faster}
\vspace{-1.0em}
\end{table}
\begin{table}[ht]
\vspace{-1.0em}
\scalebox{0.9}{\input{tables/labeled_resampling_fcos}}
\vspace{-6pt}
\centering
\caption{Powers of Labeled Resampling on FCOS.}
\label{tab:resampling_fcos}
\vspace{-1.0em}
\end{table}
\begin{table}[ht]
\vspace{-1.0em}
\scalebox{0.9}{\input{tables/confidence_per_cat}}
\vspace{-6pt}
\centering
\caption{Confidence Score of Pseudo-Labels across categories.}
\label{tab:confidence_per_cat}
\vspace{-1.0em}
\end{table}

%% file: tables/coco_standard.tex
\begin{tabular}{p{6.5cm}|c|c|c|c|c}
\toprule
& \multicolumn{4}{c|}{COCO-Standard}  &  COCO-Full \\ \cmidrule{2-5}
& 1\% & 2\% & 5\% & 10\%  & 100\% \\ 
\midrule
Supervised & 10.70 & 14.80 & 21.00 & 26.60 & 41.00\\ 
\midrule
STAC~\cite{sohn2020simple}                 & $13.97\pm0.35$ & $18.25\pm0.25$ & $24.38\pm0.12$ & $28.64\pm0.21$ & 39.48 $\xrightarrow{-0.27}$ 39.21\\

Unbiased Teacher~\cite{liu2021unbiased}    & $20.75\pm0.12$ & $24.30\pm0.07$ & $28.27\pm0.11$ & $31.50\pm0.10$ & 40.20 $\xrightarrow{+1.10}$ 41.30\\

Soft Teacher~\cite{xu2021end}              & $20.46\pm0.39$ &        -       & $30.74\pm0.08$ & $34.04\pm0.14$ & 40.90 $\xrightarrow{+3.70}$ 44.50\\ 

ACRST~\cite{zhang2022semi}                 & $\bf{26.07\pm0.46}$ & $\bf{28.69\pm0.17}$ & $31.35\pm0.13$ & $34.92\pm0.22$ & 40.20 $\xrightarrow{+2.59}$ 42.79\\

LabelMatch~\cite{chen2022label}        & $25.81\pm0.28$ &        -       & $32.70\pm0.18$ & $35.49\pm0.17$ & 40.30 $\xrightarrow{+5.00}$ 45.30\\

Unbiased Teacher v2~\cite{liu2022unbiased} & $25.40\pm0.36$ & $28.37\pm0.03$ & $31.85\pm0.09$ & $35.08\pm0.02$ & 40.90 $\xrightarrow{+3.85}$ 44.75\\ 

PseCo~\cite{li2022pseco}                   & $22.43\pm0.36$ & $27.77\pm0.18$ & $32.50\pm0.08$ & $36.06\pm0.24$ & 41.00 $\xrightarrow{+5.10}$ 46.10\\

\method~(Ours)                             & $25.06\pm0.39$ & $28.58\pm0.24$ & $\bf{33.48\pm0.13}$ & $\bf{37.16\pm0.15}$ & 41.00 $\xrightarrow{+5.20}$ \bf{46.20}\\ 
\bottomrule
\end{tabular}

% CSD\cite{jeong2019consistency}

% STAC\cite{sohn2020simple}

% ISD\cite{jeong2021interpolation}

% Unbiased Teacher\cite{liu2021unbiased}

% Humble Teacher\cite{tang2021humble}

% Multi-Phase Learning\cite{wang2021data}

% Combating Noise\cite{wang2021combating}

% Soft Teacher\cite{xu2021end}

% ISMT\cite{yang2021interactive}

% Instant-Teaching\cite{tianfzhou2021instant}

% DETReg\cite{bar2022detreg}

% Label Matching\cite{chen2022label}

% Dense Learning\cite{chen2022dense}

% SED\cite{guo2022scale}

% MUM\cite{kim2022mum}

% MA-GCP\cite{li2022semi}

% PseCo\cite{li2022pseco}

% RPL\cite{li2022rethinking}

% CST\cite{liu2022cycle}

% Unbiased Teacher v2\cite{liu2022unbiased}

% ASTOD\cite{vandeghen2022adaptive}

% Active Teacher\cite{mi2022active}

% DCST\cite{wang2022double}

% Omni-DETR\cite{wang2022omni}

% SCMT\cite{xiong2022scmt}

% ACRST\cite{zhang2022semi}

% Polishing Teacher\cite{zhang2022mind}

% S4OD\cite{zhang2022s4od}

% DDT\cite{zheng2022dual}

% Dense Teacher\cite{zhou2022dense}

%% file: tables/coco_fcos.tex
  \scalebox{0.9}{
\begin{tabular}{p{3cm}|c|c|c|c|c}
\toprule
Methods & 1\% & 2\% & 5\% & 10\%  & 100\% \\
\midrule
DSL                  & 22.0 & 25.2 & 30.9 & 36.2 & 40.2 $\xrightarrow{+3.6}$ 43.8 \\
Dense Teacher        & 22.4 & 27.2 & \bf{33.0} & 37.1 & 41.2 $\xrightarrow{+4.9}$ 46.1 \\
\method              & \bf{23.9} & \bf{27.8} & 32.8 & \bf{37.3} & 41.2 $\xrightarrow{+6.4}$ \bf{47.6} \\ 
\bottomrule
\end{tabular}
}

%% file: tables/coco_dino.tex
  \scalebox{0.9}{
\begin{tabular}{p{3cm}|c|c|c|c|c}
\toprule
Methods & 1\% & 2\% & 5\% & 10\%  & 100\%\\
\midrule
Semi-DETR  & 30.5 &   -   & \bf{40.1} & 43.5 & 48.6 $\xrightarrow{+1.8}$ 50.4\\
\method    & \bf{31.7} & \bf{34.7} & \bf{40.1} & \bf{44.6} & 50.9 $\xrightarrow{+4.3}$ \bf{55.2}\\ 
\bottomrule
\end{tabular}
}

%% file: tables/unlabel2017.tex
  \scalebox{0.9}{

\begin{tabular}{c | c | c }
 \toprule
Methods & Detectors &   AP\\
 \midrule
Soft Teacher & HTC++ & 58.2 $\xrightarrow{+1.7}$ 59.9\\
\method & DINO & 57.7 $\xrightarrow{+2.5}$ \bf{60.2}\\
\bottomrule
\end{tabular}
}

%% file: tables/voc0712.tex
\scalebox{0.9}{

\begin{tabular}{c | c | c | c}
\toprule
Detectors & Methods & AP$_{50}$ & AP$_{50:95}$ \\
 \midrule
Faster R-CNN & Supervised & 72.63 & 42.13\\ 
 \midrule 
\multirow{4}{*}{Faster R-CNN} 
& STAC & 77.45 & 44.64 \\
& Unbiased Teacher & 77.37 & 48.69 \\
& ACRST & 78.16 & 50.12 \\
& LabelMatch & 85.48 & 55.11 \\
& Unbiased Teacher V2  & 81.29 & 56.87\\ 
& \method  & \bf{85.80} & \bf{56.10} \\ 
\midrule 
\multirow{4}{*}{FCOS}  & DenseTeacher  & 79.89 & 55.87 \\
& DSL  & 80.70 & 56.80 \\ 
& \method  & \bf{84.70} & \bf{59.00} \\ 
\bottomrule 
\end{tabular}
}

%% file: tables/coco_10percent.tex
 % \tablestyle{2pt}{1.0}
  \scalebox{0.9}{

\begin{tabular}{c | c | c | c}
\toprule
 Detectors & Supervised & DetMeanTeacher & \method  \\
\midrule
RetinaNet       & 27.2 & 34.1 (+6.9) & 36.0 (+1.9) \\
CenterNet       & 27.7 & 34.9 (+7.2) & 36.7 (+1.8) \\ 
FCOS            & 26.9 & 35.9 (+9.0) & 37.3 (+1.4) \\ 
ATSS            & 28.2 & 35.5 (+7.3) & 37.0 (+1.5) \\ 
TOOD            & 29.5 & 38.2 (+8.7) & 38.9 (+0.7) \\ 
Faster R-CNN    & 26.6 & 34.7 (+8.1) & 37.2 (+2.5) \\ 
Cascade R-CNN   & 28.0 & 37.3 (+9.3) & 40.0 (+2.7)\\ 
Sparse R-CNN    & 29.3 & 36.8 (+7.5) & 38.2 (+1.4)\\ 
Deformable DETR & 31.3 & 39.3 (+8.0) & 40.5 (+1.2) \\ 
DAB DETR       & 27.5 & 33.7 (+6.2) & 36.2 (+2.5) \\ 
DINO           & 35.7 & 43.2 (+7.5) & 44.4 (+1.2)\\ 
\bottomrule
\end{tabular}
}

%% file: tables/ablation_overall.tex
 % \tablestyle{3pt}{1.0}
  \scalebox{0.9}{

\begin{tabular}{c | c |c |c c|c c c}
\toprule
\#  & Method & AP & AP$_{50}$ & AP$_{75}$ & AP$_{s}$  & AP$_{m}$  & AP$_{l}$ \\
 \midrule
1 & DetMeanTeacher      & 34.7 & 54.7 & 37.4 & 19.3 & 37.6 & 45.8 \\
2 & \#1 + Pseudo Mixup  & 35.7 & 56.0 & 39.2 & 20.3 & 38.1 & 47.2 \\
3 & \#2 + Pseudo Mosaic & 37.2 & 57.6 & 40.1 & 23.0 & 39.9 & 47.7 \\
\bottomrule
\end{tabular}
}

%% file: tables/ablation_mixup.tex
 % \tablestyle{3pt}{1.0}
  \scalebox{0.9}{

\begin{tabular}{c | c |c |c c|c c c}
\toprule
\#  & Method & AP & AP$_{50}$ & AP$_{75}$ & AP$_{s}$  & AP$_{m}$  & AP$_{l}$ \\
 \midrule
1 & DetMeanTeacher     & 34.7 & 54.7 & 37.4 & 19.3 & 37.6 & 45.8\\
2 & \#1 + GT Mixup     & 33.1 & 52.6 & 35.5 & 17.9 & 34.7 & 44.1\\
3 & \#1 + Pseudo Mixup & 35.7 & 56.0 & 39.2 & 20.3 & 38.1 & 47.2\\
\bottomrule
\end{tabular}
}

%% file: tables/ablation_mosaic.tex
 % \tablestyle{3pt}{1.0}
  \scalebox{0.9}{

\begin{tabular}{c | c |c |c c|c c c}
\toprule
\#  & Resolution & AP & AP$_{50}$ & AP$_{75}$ & AP$_{s}$  & AP$_{m}$  & AP$_{l}$ \\
 \midrule
1 & 200    & 36.1 & 56.6 & 39.2 & 22.6 & 38.3 & 46.5 \\
2 & 400    & 36.8 & 57.2 & 40.0 & \bf{23.3} & 39.4 & 47.7 \\
3 & 600    & 37.0 & 57.4 & 40.3 & 22.7 & 39.8 & 47.8 \\
4 & 800    & 37.2 & 57.5 & \bf{40.5} & 22.4 & 39.8 & 48.1 \\
5 & 1000   & 37.1 & 57.4 & 40.5 & 22.1 & \bf{40.0} & \bf{48.9} \\
6 & 200 $\sim$ 800  & 37.0 & \bf{57.7} & 40.2 & 23.0 & 39.5 & 48.0 \\
7 & 400 $\sim$ 800  & \bf{37.2} & 57.6 & 40.1 & 23.0 & 39.9 & 47.7 \\
8 & 600 $\sim$ 800  & 37.1 & 57.4 & 40.4 & 22.3 & 39.9 & 47.9 \\
\bottomrule
\end{tabular}
}

%% file: tables/labeled_resampling_faster.tex
 % \tablestyle{3pt}{1.0}
  \scalebox{0.8}{

\begin{tabular}{c |c | c |c |c c|c c c}
\toprule
\# & Method & power & mAP & person & car & toaster  & hair drier  \\
 \midrule
1 & Supervised & 0 & 26.6 & 43.8 & 34.2 & 16.8 & 0.4 \\
 \midrule
2 & MixPL      & 0 & 37.1 & 52.3 & \textbf{43.6} & 21.8 & 0.0  \\
3 & MixPL      & 0.25 & 37.1 & 52.3 & 43.2 & 28.7 & 1.5  \\
4 & MixPL      & \textbf{0.5}  & \textbf{37.2} & \textbf{52.4} & 43.1 & 26.1 & \textbf{8.1} \\
5 & MixPL      & 1.0  & 36.9 & 52.0 & 42.9 & \textbf{41.3} & 4.7  \\
\bottomrule
\end{tabular}
}

%% file: tables/labeled_resampling_fcos.tex
 % \tablestyle{3pt}{1.0}
  \scalebox{0.8}{

\begin{tabular}{c | c | c |c |c c| c c}
\toprule
\#  & Methed & Power & mAP & person & car & toaster & hair drier  \\
 \midrule
1 & Supervised & 0 & 26.9 & 44.9 & 33.4 & 18.8 & 0.0 \\
 \midrule
2 & MixPL & 0    & 37.1 & \textbf{54.2} & 43.2 & 16.4 & 1.1  \\
3 & MixPL & 0.25 & 37.1 & 54.0 & \textbf{43.5} & 34.3 & 1.4  \\
4 & MixPL & \textbf{0.5}  & \textbf{37.5} & 54.0 & 43.2 & 50.0 & 0.5 \\
5 & MixPL & 1.0  & 37.3 & 53.8 & 42.4 & \textbf{57.7} & \textbf{9.5}  \\
\bottomrule
\end{tabular}
}

%% file: tables/confidence_per_cat.tex
 % \tablestyle{3pt}{1.0}
  \scalebox{0.8}{

\begin{tabular}{c | c |c |c c|c c c}
\toprule
\# & Detectors & Loss & person & car & toaster  & hair drier  \\
 \midrule
1 & Faster R-CNN & CE Loss & 0.433  & 0.366 & 0.218 & 0.143 \\
2 & FCOS         & Focal Loss        & 0.120  & 0.117 & 0.079 & 0.057  \\
\bottomrule
\end{tabular}
}

%% file: sec/6_con.tex
% !TEX root = ../main.tex

\section{Conclusion}
This paper analyzes and solves the limitations of pseudo-labels in semi-supervised object detection frameworks. 
We first summarize a DetMeanTeacher to decouple detectors and the semi-supervised learning framework. Based on that, we observe consistent differences between pseudo-labels and ground truth in quantity, size, and categories across various detectors, and propose \method~with Labeled Resampling.
\method~alleviate the persistent negative interference of false negatives through Pseudo Mixup, and enhance the detector's performance across different sizes and categories comprehensively through Pseudo Mosaic and Label Resampling.
In addition, \method~is the first detector-agnostic framework for SSOD that unanimously and effectively improves different object detectors by unlabeled images, and sets new records of SSOD on COCO benchmarks.

% \newpage

%% file: sec/7_suppl.tex
% !TEX root = ../main.tex

\section{Cache of Pseudo-Labels}

As shown in Fig.~\ref{fig:meanteacher} of the main paper, \method~is applied after obtaining the strongly augmented unlabeled images in DetMeanTeacher.
Since \method~requires more pseudo-labeled images to mix, during each iteration, the pseudo-labeled images in the current batch are mixed with those sampled from the cache of pseudo-labels, which stores the pseudo-labels and their corresponding images in the nearest previous training iteration. When caching pseudo-labeled images, we exclude any padding from the images to save memory, and we pad these images again based on their respective shapes to make them can form a batch.
% The update of the cache follows the first-in-first-out manner. 
For example, given 1 labeled image and 4 pseudo-labeled images at each training iteration, we first sample 4 more images with their pseudo-labeles from the cache and obtain 4 mixed images by Pseudo Mixup. Then we form one mixed image using Pseudo Mosaic that randomly combines 4 of the eight images. These five mixed images are used to replace the original 4 pseudo-labeled images and are used to train the student model.

\section{The Perspective of Gradient}

 \subsection{Definition of Gradient Norm}
We adopt the definition and calculation of gradient norm in GHM~\cite{GHMLoss}.
Given a predicted bounding box, we note $p$ as the probability of the box and $p^*$ as its ground-truth label for a certain class, the binary cross entropy loss is calculated as 
\begin{equation}
\label{eq:ce1}
  L_{CE}(p,p^*) =  -p \log p - (1-p^*) \log(1-p), 
\end{equation}
where $0 \leq p \leq 1$ and $p^* \in \{0,1\}$. 
We define $x$ as the direct output of the model such that $p = \text{sigmoid}(x)$, the gradient \wrt $x$ is
\begin{equation}
\begin{aligned}
\frac{\partial L_{CE}}{\partial x} =  p - p^* .
\end{aligned}
\end{equation}
The gradient norm $g$ \wrt $x$ is thus calculated as
\begin{equation}
\label{eq:g_cls}
    g = |p - p^*| = \left\{
    \begin{aligned}
    & 1 - p  & \text{if } p^* = 1, \\
    & p & \text{if } p^* = 0,
    \end{aligned}
    \right.
\end{equation}
which represents the influence of that sample on the overall gradient. 
While the precise definition of gradient pertains to the entire parameter space, $g$ serves as a relative norm of the gradient from a sample. 
In this paper, Eq.~\ref{eq:g_cls} is used to calculate the gradient norm of positive and negative samples for RetinaNet~\cite{lin2017focal}.

\begin{figure*}[ht]
    \centering
    \includegraphics[width=1.0\linewidth]{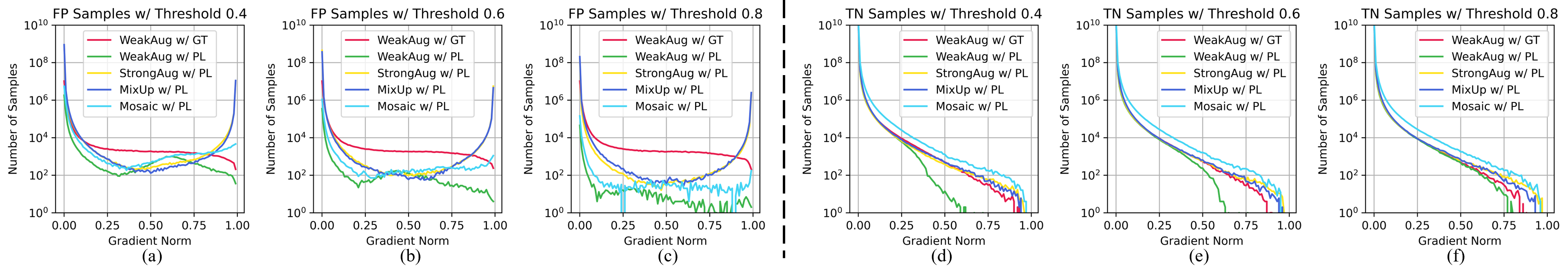}
    \caption{Gradient density for different augmentations with ground truth (GT) or pseudo labels (PL). 
    (a)-(c) represent gradient density of FP samples at various confidence thresholds; 
    (d)-(f) represent gradient density of TN samples at various confidence thresholds.}
    \label{fig:fp_tn_grad-norm}
\end{figure*}

\begin{figure*}[t]
    \centering
    \includegraphics[width=1.0\linewidth]{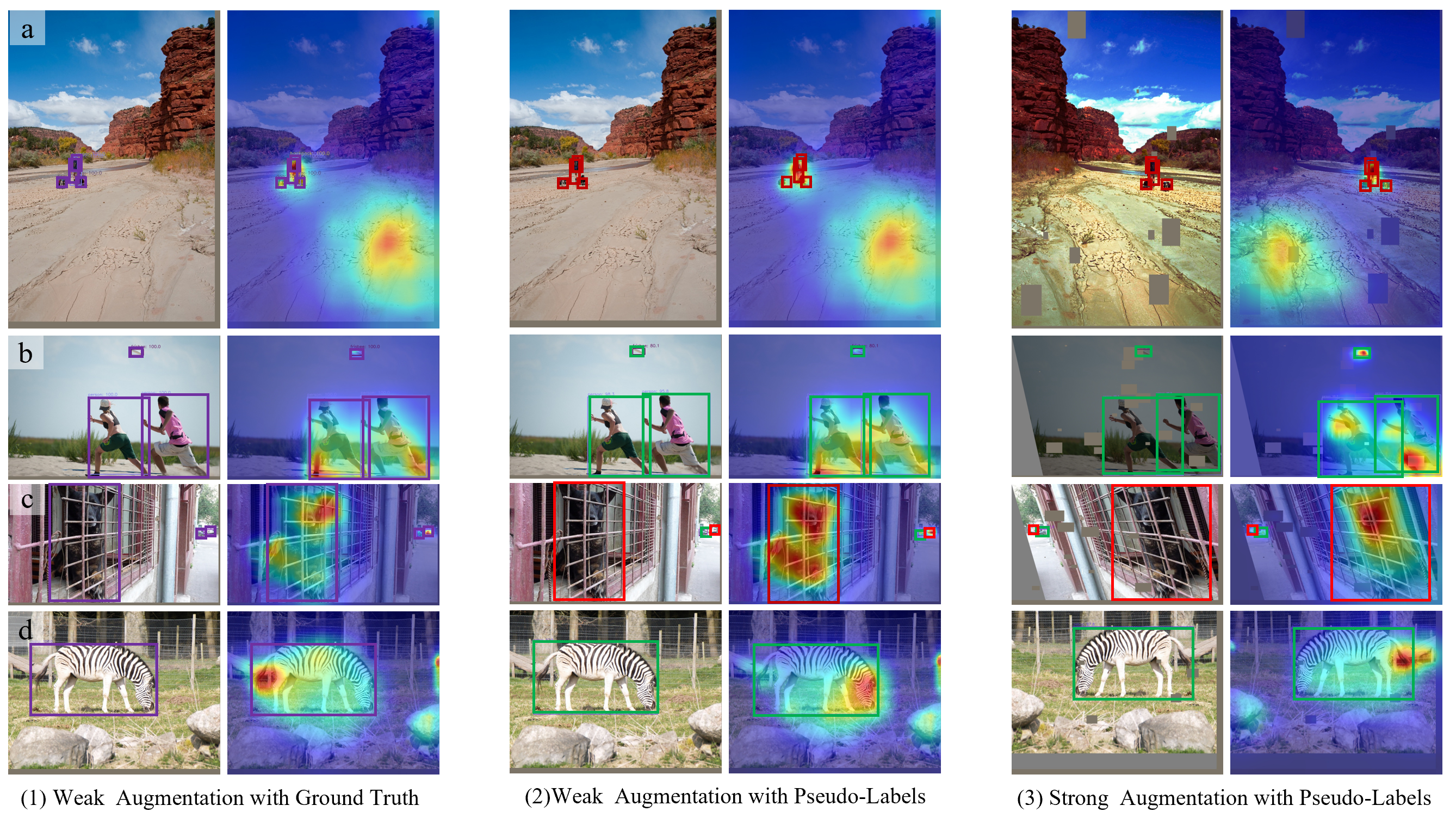}
    \caption{Grad-CAM of different augmented images with ground turth or pseudo-labels. The purple box indicates ground truth, the green box indicates the TP pseudo-labels, and the red box indicates the FN pseudo-labels.}
    \label{fig:gradnorm_gt_weak_strong}
\end{figure*}

\begin{figure*}[t]
    \centering
    \includegraphics[width=1.0\linewidth]{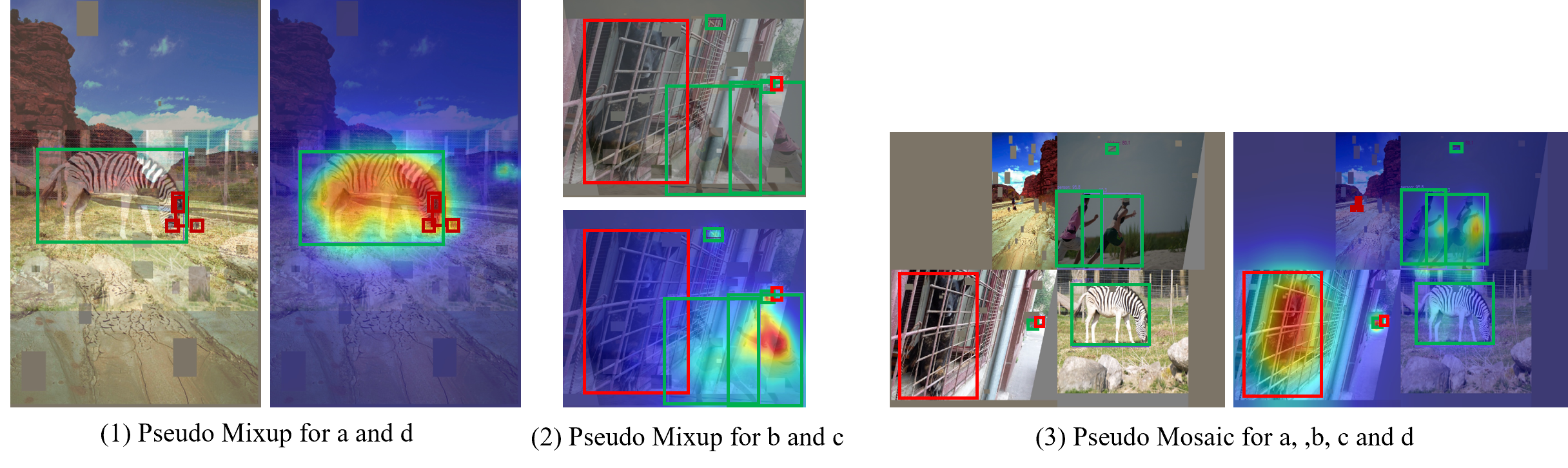}
    \caption{Grad-CAM of images with Pseudo Mixup and Pseudo Mosaic. The purple box indicates ground truth, the green box indicates the TP pseudo-labels, and the red box indicates the FN pseudo-labels.}
    \label{fig:mixup_mosaic}
\end{figure*}
\subsection{Analysis of Gradient Norm for FP and TN}

In the main text, we analyze the distribution of gradient norms for TP and FN under different data augmentations. Here we supplement the analysis for FP and TN.

As shown in Figures~\ref{fig:fp_tn_grad-norm} (a-c), strong augmentation increases the number of samples with high gradient norm for FP, which is harmful as it forces the model to learn to predict these false positive boxes.
Compared to strong augmentation, Pseudo Mixup has fewer samples with high gradient norm, which weakens the negative effect of FP, when the confidence threshold for filtering pseudo-labels is low, \ie, there are more FPs in the pseudo-labels.
However, when the confidence threshold is high, \ie, there are fewer FPs in the pseudo-labels, Pseudo Mixup contributes more low-gradient-norm samples than strong data augmentation.
Different from Pseudo Mixup, the alleviating effect of Pseudo Mosaic on FP is positively correlated with the threshold as it contributes fewer samples with gradient norms of different values.
This is because more small objects are filtered out due to their low confidence, when down-sampling the images in Pseudo Mosaic, these smaller objects will be more unrecognizable for the student model (as shown in Fig.~\ref{fig:mixup_mosaic}), which eventually reduces the impact of small-size FN.
% Furthermore, the higher the threshold, the fewer FPs but more FNs are in the pseudo-labels, and Pseudo Mosaic could add more TP in pseudo-labels to reduce the impact of FN, which implies that a higher threshold for Pseudo Mosaic might be more suitable.

We further analyze the trend of TN in Figures~\ref{fig:fp_tn_grad-norm} (d-f).
Because Pseudo Mosaic composes four images into a relatively bigger image, it always produces more true negative samples of different gradient norms in the training.
On the contrary, as Pseudo Mixup interpolates two images into a new one, the mixed image may have a different contribution from the real images, which makes it hard to recognize the objects in the mixed image, thus Pseudo Mixup contributes slightly fewer high-gradient-norm samples in TN than strong data augmentation.
We also observe that weak augmentation produces too many easy samples, \ie, contributes the fewest high-gradient-norm samples, because its augmentation level is weak.

\section{Grad-CAM with Different Augmentation}
We further apply Grad-CAM~\cite{selvaraju2017grad} to visualize and understand the behaviors of student model with the image by augmentations at different levels.
As shown in Fig.~\ref{fig:gradnorm_gt_weak_strong} (a, c), when the teacher model fails to predict a certain object and generates a false negative (FN) prediction, the student model would produce a high response under a weak augmentation at the region of the FN object(the area in red box).
This response is detrimental as it is prone to forcing the model to consider this region as background and accumulate bias during learning. Applying strong data augmentation can help mitigate this effect, as seen from the less bright heat map at the FN regions when strong augmentation is applied. 
That is, a strong augmentation can help the student avoid overfitting to FN pseudo labels as it creates a view in which missing pseudo labels are more camouflaged into the background (e.g. the dog behind the window in image c) so that the gradients from missing pseudo labels are largely suppressed. On the other hand, images b and d of Fig.~\ref{fig:gradnorm_gt_weak_strong} reveal that the gradient response in the region of TP object is more prominent when strong augmentation is applied when compared to weak data augmentation, meaning that a more difficult true prediction would accelerate training by creating larger gradient response. 

As revealed in the main manuscript and also shown in Fig.~\ref{fig:mixup_mosaic}, Pseudo Mixup functions similarly to a strong augmentation as Pseudo Mixup is effective in reducing the gradient response of FN objects (the area in red box, as shown in the right figure) and enhancing the gradient response of TP objects (the area in green box, as shown in the left figure). On the other hand, Pseudo Mosaic can increase the gradient response area of small and medium-sized TP objects, which may lead to a decrease in the response of FN objects in a specific sub-image (the upper-left sub-image), but could also introduce the response of FN objects in other sub-images (the lower-left sub-image).

\section{Implementation Details}
\subsection{Details of Different Detectors}
In semi-supervised training, we use the same optimizer and learning rate as these detectors use in the fully supervised training as shown in Table.~\ref{tab:model_semi-setting}.
Specifically, for one-stage~\cite{lin2017focal, tian2019fcos, zhang2020bridging, zhou2019objects, feng2021tood}, two-stage~\cite{ren2015faster}, and cascade~\cite{cai2019cascade} object detectors, we use SGD optimizer with a fixed learning rate of 0.01.
For query-based detection models such as Sparse R-CNN~\cite{sun2021sparse} and detection transformers~\cite{zhu2020deformable, liu2022dab, zhang2022dino}, we utilize AdamW~\cite{adamw} optimizer with the constent learning rate of $2.5e-5$, and $1e-4$, respectively.

For the confidence threshold that filters pseudo-labels, the detection models that use Cross-Entropy Loss as the classification loss function adopt a value of 0.7, while those using Focal Loss adopt a value of 0.4 except FCOS. Because the confidence of FCOS is the geometric mean of the center-ness score and the classification score, which is generally smaller, the threshold for FCOS is set to 0.3.

For the number of training iterations, due to the fast convergence speed of DINO~\cite{zhang2022dino}, we train DINO by only 90k iterations on COCO-standard to avoid overfitting, while other detection models are trained for 180k iterations.

\subsection{Details of Augmentation}
We list the implementation details of weak and strong augmentations in Table.~\ref{tab:randaugment} and Table.~\ref{tab:augmentation}.
The details generally follow the data augmentation of Soft Teacher~\cite{xu2021end} except that of RandomErasing~\cite{zhong2020random}. Considering that RandomErasing may completely erase a pseudo-labeled instance, we reduce the random size range of a single erased region and filter out pseudo-instance whose erased region ratio is greater than 70\%.

\begin{table}[t]
\centering
\caption{Semi-supervised training configuration of different detection models on COCO-standard}
\vspace{-6pt}
\scalebox{0.9}{\input{tables/model_semi-setting}}
\label{tab:model_semi-setting}
\vspace{-12pt}
\end{table}

\begin{table*}[t]
\centering
\caption{RandAugment for Detection}
\vspace{-6pt}
\scalebox{0.9}{\input{tables/randaugment}}
\label{tab:randaugment}
\vspace{-6pt}
\end{table*}

\begin{table*}[t]
\centering
\caption{Augmentation used in DetMeanTeacher}
\vspace{-6pt}
\scalebox{0.9}{\input{tables/augmentation}}
\label{tab:augmentation}
\vspace{-12pt}
\end{table*}

%% file: tables/model_semi-setting.tex
 % \tablestyle{2pt}{1.0}
  \scalebox{0.9}{

\begin{tabular}{c | c | c | c | c | c} 
 \midrule
\#  & \multirow{2}{*}{Detectors} & \multicolumn{2}{c|}{Optimizer} & \multirow{2}{*}{Threshold} & \multirow{2}{*}{Iteration} \\
\#  &  & Type & LR & & \\
 \midrule
1 & RetinaNet       & SGD & 0.01        & 0.4 & 180k\\
2 & FCOS            & SGD & 0.01        & 0.3 & 180k\\ 
3 & ATSS            & SGD & 0.01        & 0.4 & 180k\\ 
4 & CenterNet       & SGD & 0.01        & 0.4 & 180k\\ 
5 & TOOD            & SGD & 0.01        & 0.4 & 180k\\ 
6 & Faster R-CNN    & SGD & 0.01        & 0.7 & 180k\\ 
7 & Cascade R-CNN   & SGD & 0.01        & 0.7 & 180k\\ 
8 & Sparse R-CNN    & AdamW & $2.5e-5$  & 0.4 & 180k\\ 
9 & Deformable DETR & AdamW & $1e-4$    & 0.4 & 180k\\ 
10 & DAB DETR       & AdamW & $1e-4$    & 0.4 & 180k\\ 
11 & DINO           & AdamW & $1e-4$    & 0.4 & 90k\\ 
\midrule
\end{tabular}
}

%% file: tables/randaugment.tex
 % \tablestyle{2pt}{1.0}
  \scalebox{0.9}{

\begin{tabular}{c | c | c | c | p{10cm} } 
\hline
\#  & Space & Type & Magnitude &  Description \\
\hline
1 & \multirow{14}{*}{Color}  & AutoContrast  & - & \makecell*[l]{Maximize the the image contrast, by making the darkest pixel black and \\ lightest pixel white.} \\
\cline{3-5}
2 &                        & Equalize  & - & \makecell*[l]{Equalize the image histogram.}\\
\cline{3-5}
3 &                        & Solarize & [0, 256] & \makecell*[l]{Invert all pixels above a threshold value of \emph{magnitude}.}\\
\cline{3-5}
4 &                        & Posterize  & [4, 8] & \makecell*[l]{Reduce the number of bits for each pixel to \emph{magnitude} bits.}\\
\cline{3-5}
5 &                        & Contrast  & [0.1, 1.9] & \makecell*[l]{Control the contrast of the image. A \emph{magnitude}=0 gives a gray image, \\ whereas \emph{magnitude}=1 gives the original image.}\\
\cline{3-5}
6 &                        & Color  & [0.1, 1.9] & \makecell*[l]{Adjust the color balance of the image, in a manner similar to the controls \\ on a colour TV set. A \emph{magnitude}=0 gives a black \& white image, \\ whereas \emph{magnitude}=1 gives the original image.}\\
\cline{3-5}
7 &                        & Brightness  & [0.1, 1.9] & \makecell*[l]{Adjust the brightness of the image. A \emph{magnitude}=0 gives a black image, \\ whereas \emph{magnitude}=1 gives the original image.}\\
\cline{3-5}
8 &                        & Sharpness  & [0.1, 1.9] & \makecell*[l]{Adjust the sharpness of the image. A \emph{magnitude}=0 gives a blurred image, \\ whereas \emph{magnitude}=1 gives the original image.}\\
\hline
9 & \multirow{3}{*}{Geometric}  & ShearX(Y) & [-0.3, 0.3] & \makecell*[l]{Shear the image and bounding boxes along the horizontal (vertical) \\ axis with \emph{magnitude} degrees.} \\
\cline{3-5}
10 &                        & TranslateX(Y)  & [-0.1, 0.1] & \makecell*[l]{Translate the image and bounding boxes  in the horizontal (vertical) \\ direction with rate \emph{magnitude}.} \\
\cline{3-5}
11 &                        & Rotate & [-30, 30] & \makecell*[l]{Rotate the image and bounding boxes \emph{magnitude} degrees.} \\
\hline
\end{tabular}
}

%% file: tables/augmentation.tex
 % \tablestyle{2pt}{1.0}
  \scalebox{0.9}{

\begin{tabular}{c | c | c | c | p{9cm}} 
\hline
\#  & Pipline & Type & Parameter &  Description\\
\hline
1 & \multirow{6}{*}{Labeled}  & RandomResize & scale = [(1333, 400), (1333, 1200)] & \makecell*[l]{Randomly scale the image while maintaining aspect ratio, \\ where the longer side is less than 1333 and the shorter side \\ is between 400 and 1200.}\\
\cline{3-5}
2 &                        & RandomFlip & prob=0.5 & Randomly flip the image horizontally with a probability of 0.5.\\
\cline{3-5}
3 &                        & RandAugment & aug\_space=color\_space, aug\_num=1 & \makecell*[l]{Randomly select 1 color transformation from the pre-defined set \\ of RandAugment color transformations to apply to the image.}\\
\hline
4 & \multirow{2}{*}{Weak}  & RandomResize & scale = [(1333, 400), (1333, 1200)] & \makecell*[l]{Same as \#1.} \\
\cline{3-5}
5 &                        & RandomFlip & prob=0.5 & \makecell*[l]{Same as \#2.} \\
\hline
6 & \multirow{8}{*}{Strong} & RandomResize & scale = [(1333, 400), (1333, 1200)] & \makecell*[l]{Same as \#1.}\\ 
\cline{3-5}
7 &                        & RandomFlip & prob=0.5 & \makecell*[l]{Same as \#2.}\\
\cline{3-5}
8 &                        & RandAugment & aug\_space=color\_space, aug\_num=1 & \makecell*[l]{Same as \#3.}\\
\cline{3-5}
9 &                        & RandAugment & aug\_space=geometric, aug\_num=1 & \makecell*[l]{Randomly select 1 geometric transformation from the pre-defined \\ set of RandAugment color transformations to \\ apply to the image.}\\
\cline{3-5}
10 &                       & RandErasing & patches=(1, 20), ratio=(0, 0.1), thr=0.7 & \makecell*[l]{Randomly erase regions with side length ratio between 0 and 10\%, \\ for 1 to 20 regions. Also, filter out pseudo-labels where the erased \\ area exceeds 70\% of the total area.}\\
\hline
\end{tabular}
}

%% file: main.bbl
\begin{thebibliography}{52}
\providecommand{\natexlab}[1]{#1}
\providecommand{\url}[1]{\texttt{#1}}
\expandafter\ifx\csname urlstyle\endcsname\relax
  \providecommand{\doi}[1]{doi: #1}\else
  \providecommand{\doi}{doi: \begingroup \urlstyle{rm}\Url}\fi

\bibitem[Berthelot et~al.(2019)Berthelot, Carlini, Goodfellow, Papernot, Oliver, and Raffel]{berthelot2019mixmatch}
David Berthelot, Nicholas Carlini, Ian Goodfellow, Nicolas Papernot, Avital Oliver, and Colin~A Raffel.
\newblock Mixmatch: A holistic approach to semi-supervised learning.
\newblock In \emph{NeurIPS}, 2019.

\bibitem[Bochkovskiy et~al.(2020)Bochkovskiy, Wang, and Liao]{yolov4}
Alexey Bochkovskiy, Chien{-}Yao Wang, and Hong{-}Yuan~Mark Liao.
\newblock Yolov4: Optimal speed and accuracy of object detection.
\newblock \emph{CoRR}, abs/2004.10934, 2020.

\bibitem[Cai and Vasconcelos(2019)]{cai2019cascade}
Zhaowei Cai and Nuno Vasconcelos.
\newblock Cascade r-cnn: high quality object detection and instance segmentation.
\newblock \emph{IEEE TPAMI}, 2019.

\bibitem[Carion et~al.(2020)Carion, Massa, Synnaeve, Usunier, Kirillov, and Zagoruyko]{detr}
Nicolas Carion, Francisco Massa, Gabriel Synnaeve, Nicolas Usunier, Alexander Kirillov, and Sergey Zagoruyko.
\newblock End-to-end object detection with transformers.
\newblock In \emph{ECCV}, 2020.

\bibitem[Chen et~al.(2022{\natexlab{a}})Chen, Chen, Yang, Xuan, Song, Xie, Pu, Song, and Zhuang]{chen2022label}
Binbin Chen, Weijie Chen, Shicai Yang, Yunyi Xuan, Jie Song, Di Xie, Shiliang Pu, Mingli Song, and Yueting Zhuang.
\newblock Label matching semi-supervised object detection.
\newblock In \emph{CVPR}, 2022{\natexlab{a}}.

\bibitem[Chen et~al.(2022{\natexlab{b}})Chen, Li, Chen, Wang, Zhang, and Hua]{chen2022dense}
Binghui Chen, Pengyu Li, Xiang Chen, Biao Wang, Lei Zhang, and Xian-Sheng Hua.
\newblock Dense learning based semi-supervised object detection.
\newblock In \emph{CVPR}, 2022{\natexlab{b}}.

\bibitem[Chen et~al.(2019{\natexlab{a}})Chen, Pang, Wang, Xiong, Li, Sun, Feng, Liu, Shi, Ouyang, Loy, and Lin]{chen2019htc}
Kai Chen, Jiangmiao Pang, Jiaqi Wang, Yu Xiong, Xiaoxiao Li, Shuyang Sun, Wansen Feng, Ziwei Liu, Jianping Shi, Wanli Ouyang, Chen~Change Loy, and Dahua Lin.
\newblock Hybrid task cascade for instance segmentation.
\newblock In \emph{CVPR}, 2019{\natexlab{a}}.

\bibitem[Chen et~al.(2019{\natexlab{b}})Chen, Wang, Pang, Cao, Xiong, Li, Sun, Feng, Liu, Xu, Zhang, Cheng, Zhu, Cheng, Zhao, Li, Lu, Zhu, Wu, Dai, Wang, Shi, Ouyang, Loy, and Lin]{mmdetection}
Kai Chen, Jiaqi Wang, Jiangmiao Pang, Yuhang Cao, Yu Xiong, Xiaoxiao Li, Shuyang Sun, Wansen Feng, Ziwei Liu, Jiarui Xu, Zheng Zhang, Dazhi Cheng, Chenchen Zhu, Tianheng Cheng, Qijie Zhao, Buyu Li, Xin Lu, Rui Zhu, Yue Wu, Jifeng Dai, Jingdong Wang, Jianping Shi, Wanli Ouyang, Chen~Change Loy, and Dahua Lin.
\newblock {MMDetection}: Open mmlab detection toolbox and benchmark.
\newblock \emph{arXiv preprint arXiv:1906.07155}, 2019{\natexlab{b}}.

\bibitem[DeVries and Taylor(2017)]{devries2017improved}
Terrance DeVries and Graham~W Taylor.
\newblock Improved regularization of convolutional neural networks with cutout.
\newblock \emph{arXiv preprint arXiv:1708.04552}, 2017.

\bibitem[Everingham et~al.(2010)Everingham, Van~Gool, Williams, Winn, and Zisserman]{everingham2010pascal}
Mark Everingham, Luc Van~Gool, Christopher~KI Williams, John Winn, and Andrew Zisserman.
\newblock The pascal visual object classes (voc) challenge.
\newblock \emph{IJCV}, 2010.

\bibitem[Feng et~al.(2021)Feng, Zhong, Gao, Scott, and Huang]{feng2021tood}
Chengjian Feng, Yujie Zhong, Yu Gao, Matthew~R Scott, and Weilin Huang.
\newblock Tood: Task-aligned one-stage object detection.
\newblock In \emph{2021 IEEE/CVF International Conference on Computer Vision (ICCV)}, pages 3490--3499. IEEE Computer Society, 2021.

\bibitem[Ge et~al.(2021)Ge, Liu, Wang, Li, and Sun]{yolox}
Zheng Ge, Songtao Liu, Feng Wang, Zeming Li, and Jian Sun.
\newblock {YOLOX:} exceeding {YOLO} series in 2021.
\newblock \emph{CoRR}, abs/2107.08430, 2021.

\bibitem[Ghiasi et~al.(2019)Ghiasi, Lin, and Le]{nasfpn}
Golnaz Ghiasi, Tsung{-}Yi Lin, and Quoc~V. Le.
\newblock {NAS-FPN: Learning} scalable feature pyramid architecture for object detection.
\newblock In \emph{CVPR}, 2019.

\bibitem[Jeong et~al.(2019)Jeong, Lee, Kim, and Kwak]{jeong2019consistency}
Jisoo Jeong, Seungeui Lee, Jeesoo Kim, and Nojun Kwak.
\newblock Consistency-based semi-supervised learning for object detection.
\newblock In \emph{NeurIPS}, 2019.

\bibitem[Jeong et~al.(2021)Jeong, Verma, Hyun, Kannala, and Kwak]{jeong2021interpolation}
Jisoo Jeong, Vikas Verma, Minsung Hyun, Juho Kannala, and Nojun Kwak.
\newblock Interpolation-based semi-supervised learning for object detection.
\newblock In \emph{CVPR}, 2021.

\bibitem[Laine and Aila(2016)]{laine2016temporal}
Samuli Laine and Timo Aila.
\newblock Temporal ensembling for semi-supervised learning.
\newblock \emph{arXiv preprint arXiv:1610.02242}, 2016.

\bibitem[Lee et~al.(2013)]{lee2013pseudo}
Dong-Hyun Lee et~al.
\newblock Pseudo-label: The simple and efficient semi-supervised learning method for deep neural networks.
\newblock In \emph{Workshop on challenges in representation learning, ICML}, 2013.

\bibitem[Li et~al.(2019)Li, Liu, and Wang]{GHMLoss}
Buyu Li, Yu Liu, and Xiaogang Wang.
\newblock Gradient harmonized single-stage detector.
\newblock In \emph{AAAI}, 2019.

\bibitem[Li et~al.(2022)Li, Li, Wang, Wu, Liang, and Zhang]{li2022pseco}
Gang Li, Xiang Li, Yujie Wang, Yichao Wu, Ding Liang, and Shanshan Zhang.
\newblock Pseco: Pseudo labeling and consistency training for semi-supervised object detection.
\newblock In \emph{ECCV}. Springer, 2022.

\bibitem[Lin et~al.(2017{\natexlab{a}})Lin, Doll{\'{a}}r, Girshick, He, Hariharan, and Belongie]{lin2017fpn}
Tsung{-}Yi Lin, Piotr Doll{\'{a}}r, Ross~B. Girshick, Kaiming He, Bharath Hariharan, and Serge~J. Belongie.
\newblock Feature pyramid networks for object detection.
\newblock In \emph{ICCV}, 2017{\natexlab{a}}.

\bibitem[Lin et~al.(2014)Lin, Maire, Belongie, Hays, Perona, Ramanan, Doll{\'a}r, and Zitnick]{coco}
Tsung-Yi Lin, Michael Maire, Serge Belongie, James Hays, Pietro Perona, Deva Ramanan, Piotr Doll{\'a}r, and C~Lawrence Zitnick.
\newblock Microsoft coco: Common objects in context.
\newblock In \emph{European conference on computer vision}, pages 740--755. Springer, 2014.

\bibitem[Lin et~al.(2017{\natexlab{b}})Lin, Goyal, Girshick, He, and Doll{\'a}r]{lin2017focal}
Tsung-Yi Lin, Priya Goyal, Ross Girshick, Kaiming He, and Piotr Doll{\'a}r.
\newblock Focal loss for dense object detection.
\newblock In \emph{ICCV}, 2017{\natexlab{b}}.

\bibitem[Liu et~al.(2022{\natexlab{a}})Liu, Li, Zhang, Yang, Qi, Su, Zhu, and Zhang]{liu2022dab}
Shilong Liu, Feng Li, Hao Zhang, Xiao Yang, Xianbiao Qi, Hang Su, Jun Zhu, and Lei Zhang.
\newblock Dab-detr: Dynamic anchor boxes are better queries for detr.
\newblock \emph{arXiv preprint arXiv:2201.12329}, 2022{\natexlab{a}}.

\bibitem[Liu et~al.(2021{\natexlab{a}})Liu, Ma, He, Kuo, Chen, Zhang, Wu, Kira, and Vajda]{liu2021unbiased}
Yen-Cheng Liu, Chih-Yao Ma, Zijian He, Chia-Wen Kuo, Kan Chen, Peizhao Zhang, Bichen Wu, Zsolt Kira, and Peter Vajda.
\newblock Unbiased teacher for semi-supervised object detection.
\newblock \emph{arXiv preprint arXiv:2102.09480}, 2021{\natexlab{a}}.

\bibitem[Liu et~al.(2022{\natexlab{b}})Liu, Ma, and Kira]{liu2022unbiased}
Yen-Cheng Liu, Chih-Yao Ma, and Zsolt Kira.
\newblock Unbiased teacher v2: Semi-supervised object detection for anchor-free and anchor-based detectors.
\newblock In \emph{CVPR}, 2022{\natexlab{b}}.

\bibitem[Liu et~al.(2021{\natexlab{b}})Liu, Lin, Cao, Hu, Wei, Zhang, Lin, and Guo]{swin}
Ze Liu, Yutong Lin, Yue Cao, Han Hu, Yixuan Wei, Zheng Zhang, Stephen Lin, and Baining Guo.
\newblock Swin transformer: Hierarchical vision transformer using shifted windows.
\newblock In \emph{ICCV}, 2021{\natexlab{b}}.

\bibitem[Loshchilov and Hutter(2019)]{adamw}
Ilya Loshchilov and Frank Hutter.
\newblock Decoupled weight decay regularization.
\newblock In \emph{ICLR}. OpenReview.net, 2019.

\bibitem[Lyu et~al.(2022)Lyu, Zhang, Huang, Zhou, Wang, Liu, Zhang, and Chen]{rtmdet}
Chengqi Lyu, Wenwei Zhang, Haian Huang, Yue Zhou, Yudong Wang, Yanyi Liu, Shilong Zhang, and Kai Chen.
\newblock {RTMDet: An} empirical study of designing real-time object detectors.
\newblock \emph{CoRR}, abs/2212.07784, 2022.

\bibitem[Ren et~al.(2015)Ren, He, Girshick, and Sun]{ren2015faster}
Shaoqing Ren, Kaiming He, Ross Girshick, and Jian Sun.
\newblock Faster r-cnn: Towards real-time object detection with region proposal networks.
\newblock In \emph{NeurIPS}, 2015.

\bibitem[Selvaraju et~al.(2017)Selvaraju, Cogswell, Das, Vedantam, Parikh, and Batra]{selvaraju2017grad}
Ramprasaath~R Selvaraju, Michael Cogswell, Abhishek Das, Ramakrishna Vedantam, Devi Parikh, and Dhruv Batra.
\newblock Grad-cam: Visual explanations from deep networks via gradient-based localization.
\newblock In \emph{Proceedings of the IEEE international conference on computer vision}, pages 618--626, 2017.

\bibitem[Sohn et~al.(2020{\natexlab{a}})Sohn, Berthelot, Carlini, Zhang, Zhang, Raffel, Cubuk, Kurakin, and Li]{sohn2020fixmatch}
Kihyuk Sohn, David Berthelot, Nicholas Carlini, Zizhao Zhang, Han Zhang, Colin~A Raffel, Ekin~Dogus Cubuk, Alexey Kurakin, and Chun-Liang Li.
\newblock Fixmatch: Simplifying semi-supervised learning with consistency and confidence.
\newblock In \emph{NeurIPS}, 2020{\natexlab{a}}.

\bibitem[Sohn et~al.(2020{\natexlab{b}})Sohn, Zhang, Li, Zhang, Lee, and Pfister]{sohn2020simple}
Kihyuk Sohn, Zizhao Zhang, Chun-Liang Li, Han Zhang, Chen-Yu Lee, and Tomas Pfister.
\newblock A simple semi-supervised learning framework for object detection.
\newblock \emph{arXiv preprint arXiv:2005.04757}, 2020{\natexlab{b}}.

\bibitem[Sun et~al.(2021)Sun, Zhang, Jiang, Kong, Xu, Zhan, Tomizuka, Li, Yuan, Wang, et~al.]{sun2021sparse}
Peize Sun, Rufeng Zhang, Yi Jiang, Tao Kong, Chenfeng Xu, Wei Zhan, Masayoshi Tomizuka, Lei Li, Zehuan Yuan, Changhu Wang, et~al.
\newblock Sparse r-cnn: End-to-end object detection with learnable proposals.
\newblock In \emph{CVPR}, 2021.

\bibitem[Tarvainen and Valpola(2017)]{tarvainen2017mean}
Antti Tarvainen and Harri Valpola.
\newblock Mean teachers are better role models: Weight-averaged consistency targets improve semi-supervised deep learning results.
\newblock \emph{NeurIPS}, 2017.

\bibitem[Tian et~al.(2019)Tian, Shen, Chen, and He]{tian2019fcos}
Zhi Tian, Chunhua Shen, Hao Chen, and Tong He.
\newblock Fcos: Fully convolutional one-stage object detection.
\newblock In \emph{CVPR}, 2019.

\bibitem[Wang et~al.(2022)Wang, Chen, Heng, Hou, Savvides, Shinozaki, Raj, Wu, and Wang]{wang2022freematch}
Yidong Wang, Hao Chen, Qiang Heng, Wenxin Hou, Marios Savvides, Takahiro Shinozaki, Bhiksha Raj, Zhen Wu, and Jindong Wang.
\newblock Freematch: Self-adaptive thresholding for semi-supervised learning.
\newblock \emph{arXiv preprint arXiv:2205.07246}, 2022.

\bibitem[Wu et~al.(2019)Wu, Kirillov, Massa, Lo, and Girshick]{wu2019detectron2}
Yuxin Wu, Alexander Kirillov, Francisco Massa, Wan-Yen Lo, and Ross Girshick.
\newblock Detectron2.
\newblock \url{https://github.com/facebookresearch/detectron2}, 2019.

\bibitem[Xie et~al.(2020{\natexlab{a}})Xie, Dai, Hovy, Luong, and Le]{xie2020unsupervised}
Qizhe Xie, Zihang Dai, Eduard Hovy, Thang Luong, and Quoc Le.
\newblock Unsupervised data augmentation for consistency training.
\newblock In \emph{NeurIPS}, 2020{\natexlab{a}}.

\bibitem[Xie et~al.(2020{\natexlab{b}})Xie, Luong, Hovy, and Le]{xie2020self}
Qizhe Xie, Minh-Thang Luong, Eduard Hovy, and Quoc~V Le.
\newblock Self-training with noisy student improves imagenet classification.
\newblock In \emph{CVPR}, 2020{\natexlab{b}}.

\bibitem[Xu et~al.(2021)Xu, Zhang, Hu, Wang, Wang, Wei, Bai, and Liu]{xu2021end}
Mengde Xu, Zheng Zhang, Han Hu, Jianfeng Wang, Lijuan Wang, Fangyun Wei, Xiang Bai, and Zicheng Liu.
\newblock End-to-end semi-supervised object detection with soft teacher.
\newblock In \emph{CVPR}, 2021.

\bibitem[Zhang et~al.(2021)Zhang, Wang, Hou, Wu, Wang, Okumura, and Shinozaki]{zhang2021flexmatch}
Bowen Zhang, Yidong Wang, Wenxin Hou, Hao Wu, Jindong Wang, Manabu Okumura, and Takahiro Shinozaki.
\newblock Flexmatch: Boosting semi-supervised learning with curriculum pseudo labeling.
\newblock In \emph{NeurIPS}, pages 18408--18419, 2021.

\bibitem[Zhang et~al.(2022{\natexlab{a}})Zhang, Pan, and Wang]{zhang2022semi}
Fangyuan Zhang, Tianxiang Pan, and Bin Wang.
\newblock Semi-supervised object detection with adaptive class-rebalancing self-training.
\newblock In \emph{AAAI}, 2022{\natexlab{a}}.

\bibitem[Zhang et~al.(2017)Zhang, Cisse, Dauphin, and Lopez-Paz]{zhang2017mixup}
Hongyi Zhang, Moustapha Cisse, Yann~N Dauphin, and David Lopez-Paz.
\newblock mixup: Beyond empirical risk minimization.
\newblock \emph{arXiv preprint arXiv:1710.09412}, 2017.

\bibitem[Zhang et~al.(2022{\natexlab{b}})Zhang, Li, Liu, Zhang, Su, Zhu, Ni, and Shum]{zhang2022dino}
Hao Zhang, Feng Li, Shilong Liu, Lei Zhang, Hang Su, Jun Zhu, Lionel Ni, and Harry Shum.
\newblock Dino: Detr with improved denoising anchor boxes for end-to-end object detection.
\newblock In \emph{ICLR}, 2022{\natexlab{b}}.

\bibitem[Zhang et~al.(2023)Zhang, Lin, Zhang, Wang, Tan, Han, Ding, Wang, and Li]{zhang2023semi}
Jiacheng Zhang, Xiangru Lin, Wei Zhang, Kuo Wang, Xiao Tan, Junyu Han, Errui Ding, Jingdong Wang, and Guanbin Li.
\newblock Semi-detr: Semi-supervised object detection with detection transformers.
\newblock In \emph{Proceedings of the IEEE/CVF Conference on Computer Vision and Pattern Recognition}, pages 23809--23818, 2023.

\bibitem[Zhang et~al.(2020)Zhang, Chi, Yao, Lei, and Li]{zhang2020bridging}
Shifeng Zhang, Cheng Chi, Yongqiang Yao, Zhen Lei, and Stan~Z Li.
\newblock Bridging the gap between anchor-based and anchor-free detection via adaptive training sample selection.
\newblock In \emph{CVPR}, 2020.

\bibitem[Zheng et~al.(2022)Zheng, You, Huang, Wang, Qian, and Xu]{zheng2022simmatch}
Mingkai Zheng, Shan You, Lang Huang, Fei Wang, Chen Qian, and Chang Xu.
\newblock Simmatch: Semi-supervised learning with similarity matching.
\newblock In \emph{CVPR}, 2022.

\bibitem[Zhong et~al.(2020)Zhong, Zheng, Kang, Li, and Yang]{zhong2020random}
Zhun Zhong, Liang Zheng, Guoliang Kang, Shaozi Li, and Yi Yang.
\newblock Random erasing data augmentation.
\newblock In \emph{AAAI}, 2020.

\bibitem[Zhou et~al.(2022)Zhou, Ge, Liu, Mao, Li, Yu, and Sun]{zhou2022dense}
Hongyu Zhou, Zheng Ge, Songtao Liu, Weixin Mao, Zeming Li, Haiyan Yu, and Jian Sun.
\newblock Dense teacher: Dense pseudo-labels for semi-supervised object detection.
\newblock In \emph{ECCV}. Springer, 2022.

\bibitem[Zhou et~al.(2021)Zhou, Yu, Wang, Qian, and Li]{tianfzhou2021instant}
Qiang Zhou, Chaohui Yu, Zhibin Wang, Qi Qian, and Hao Li.
\newblock Instant-teaching: An end-to-end semi-supervised object detection framework.
\newblock In \emph{CVPR}, 2021.

\bibitem[Zhou et~al.(2019)Zhou, Wang, and Kr{\"a}henb{\"u}hl]{zhou2019objects}
Xingyi Zhou, Dequan Wang, and Philipp Kr{\"a}henb{\"u}hl.
\newblock Objects as points.
\newblock \emph{arXiv preprint arXiv:1904.07850}, 2019.

\bibitem[Zhu et~al.(2020)Zhu, Su, Lu, Li, Wang, and Dai]{zhu2020deformable}
Xizhou Zhu, Weijie Su, Lewei Lu, Bin Li, Xiaogang Wang, and Jifeng Dai.
\newblock Deformable detr: Deformable transformers for end-to-end object detection.
\newblock \emph{arXiv preprint arXiv:2010.04159}, 2020.

\end{thebibliography}
